\documentclass[letterpaper, 10 pt, conference]{ieeeconf}

\IEEEoverridecommandlockouts           
 
\usepackage[pdftex]{graphicx}
\usepackage{amssymb}
\usepackage{amsmath}
\usepackage{bm}
\usepackage{cite}
\usepackage{epsfig}
\usepackage{caption}
\usepackage{color}
\usepackage{psfrag}
\usepackage{cases}
\usepackage{acronym}
\usepackage{setspace}
\usepackage{times}
\usepackage{latexsym}
\usepackage{dblfloatfix}
\usepackage{subcaption}
\usepackage{metalogo}
\usepackage{tabularx}
\usepackage{array}
\usepackage{multirow}
\usepackage{hyperref}
\hypersetup{
    colorlinks=true,
    linkcolor=blue,
    filecolor=blue,      
    urlcolor=blue,
    citecolor=cyan,
}

\title{PASS3D: Precise and Accelerated Semantic Segmentation for 3D Point Cloud}

\author{Xin Kong$^{1}$, Guangyao Zhai$^{1}$, Baoquan Zhong$^{1}$ and Yong Liu$^{1,2}$\\
		\thanks{$^{1}$Xin Kong, Guangyao Zhai and Baoquan Zhong are with the Institute of Cyber-Systems and Control, Zhejiang University, Zhejiang, 310027, China.}%
		\thanks{$^{2}$Yong Liu is with the State Key Laboratory of Industrial Control Technology and Institute of Cyber-Systems and Control, Zhejiang University, Zhejiang, 310027, China (Yong Liu is the corresponding author, email: yongliu@iipc.zju.edu.cn).}%
}

\begin{document}

\maketitle
\thispagestyle{empty}
\pagestyle{empty}

\begin{abstract}
In this paper, we propose PASS3D to achieve point-wise semantic segmentation for 3D point cloud. Our framework combines the efficiency of traditional geometric methods with robustness of deep learning methods, consisting of two stages: At stage-1, our accelerated cluster proposal algorithm will generate refined cluster proposals by segmenting point clouds without ground, capable of generating less redundant proposals with higher recall in an extremely short time; stage-2 we will amplify and further process these proposals by a neural network to estimate semantic label for each point and meanwhile propose a novel data augmentation method to enhance the network's recognition capability for all categories especially for non-rigid objects. Evaluated on KITTI raw dataset, PASS3D stands out against the state-of-the-art on some results, making itself competent to 3D perception in autonomous driving system. Our source code will be open-sourced. A video demonstration is available at \url{https://www.youtube.com/watch?v=cukEqDuP_Qw}.
   
\end{abstract}

\section{Introduction}

Autonomous driving, a promising and accessible technology, gains more and more research attention. 3D LiDAR-based perception is one of the significant technical solutions in autonomous driving. Although 3D LiDAR scanner can provide distance measurements directly, and generate 3D point clouds to capture the geometrical structure of the scene, it is still quite challenging to segment the point clouds semantically due to the missing of the texture information. The perception of 3D scene requires semantic segmentation on point clouds, which is still an unsolved problem.

\begin{figure}
    \centering
    \scalebox{1}{
    \begin{subfigure}[b]{0.48\columnwidth}
		\includegraphics[width=\textwidth]{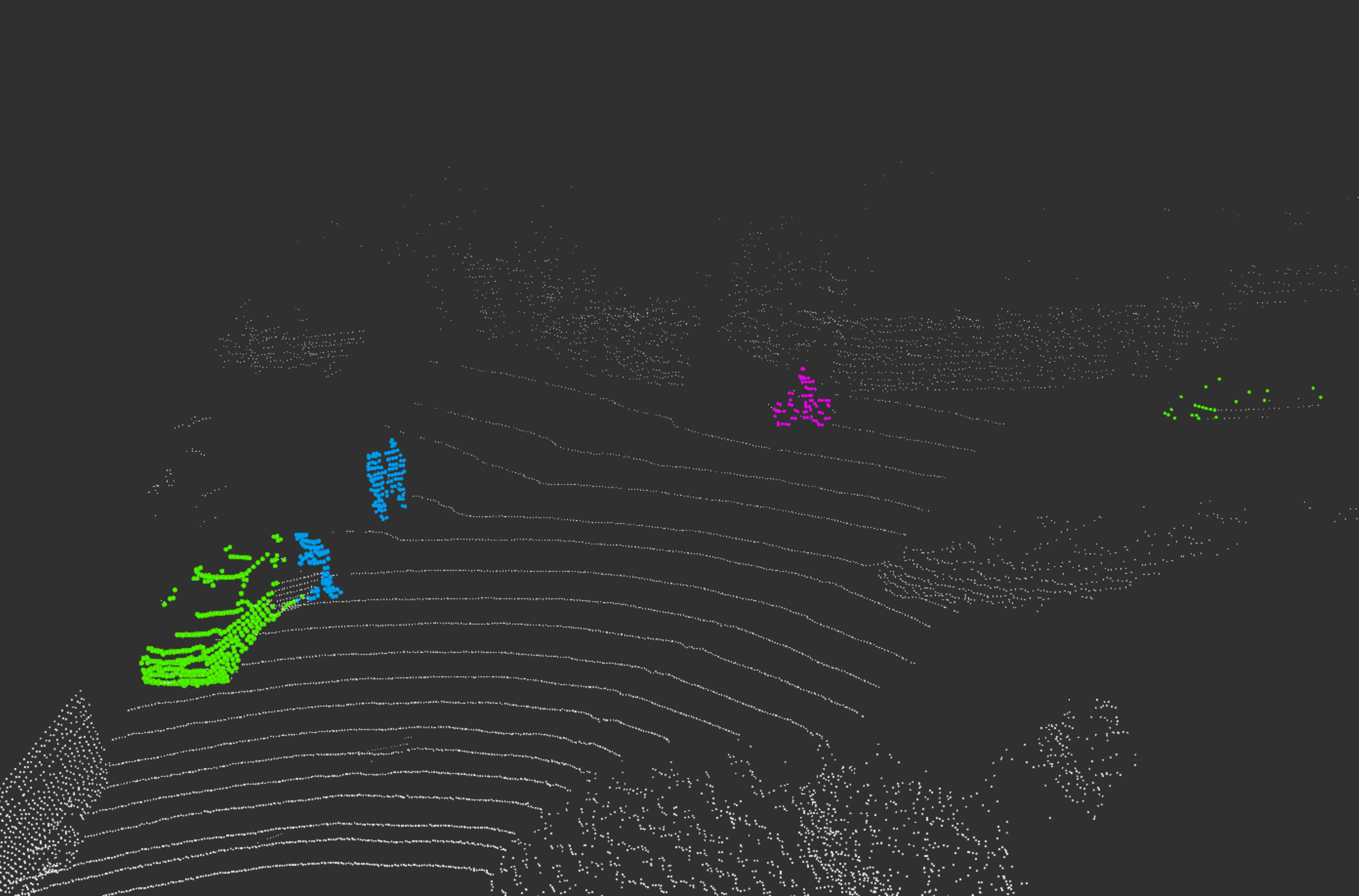}
        \caption{Ground truth}
    \end{subfigure}
    \begin{subfigure}[b]{0.48\columnwidth}
    	\includegraphics[width=\textwidth]{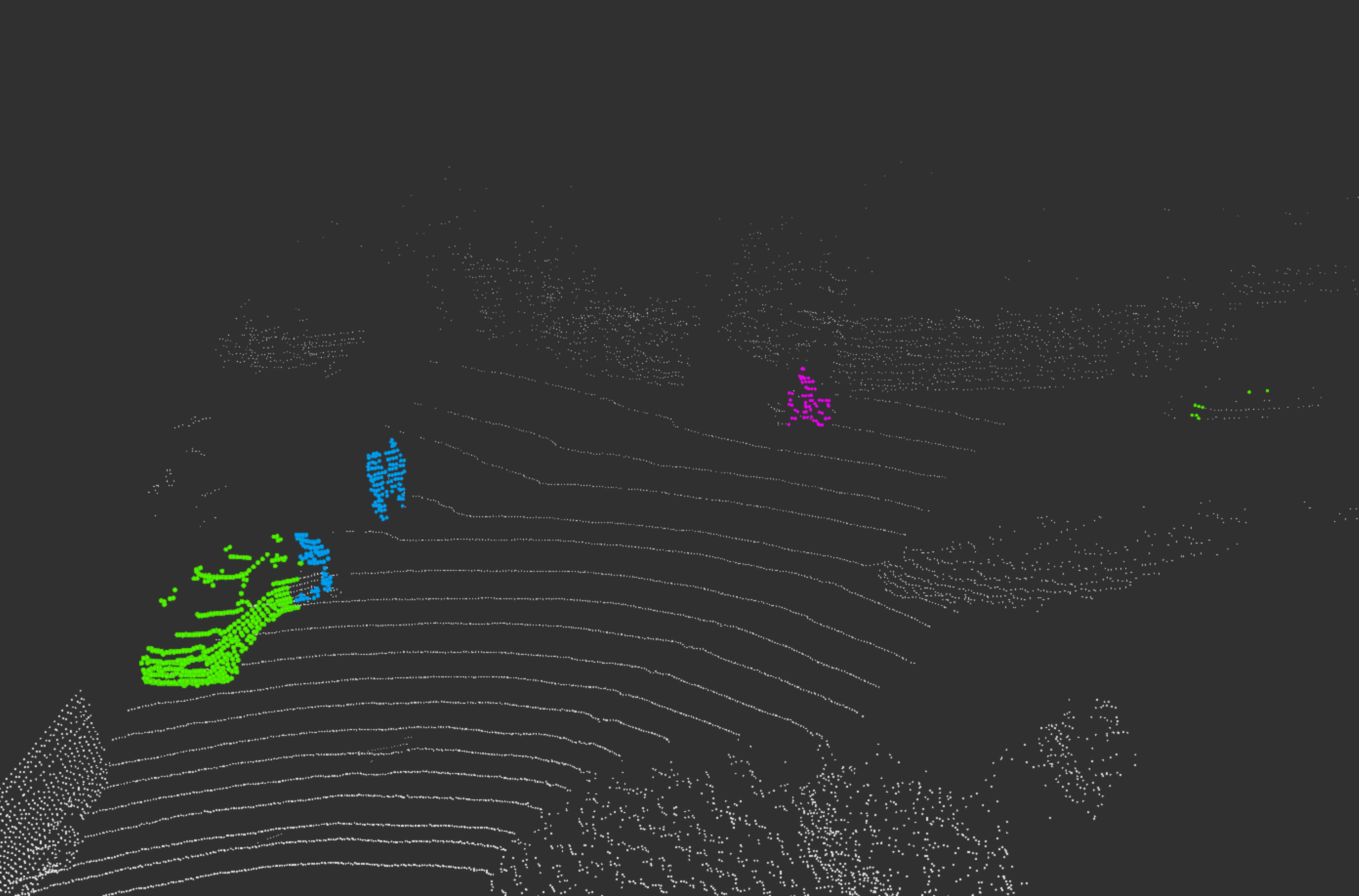}						\caption{Predicted result} 								
    \end{subfigure}
    }
    \caption{An example of PASS3D results. Our predicted result is on the right and the ground truth is on the left. `Car' is annotated in green, `Pedestrian' in blue and `Cyclist' in pink. (Best viewed with zoom-in)}
    \vspace{-3mm}
    \label{fig:example} 	
\end{figure}

Our work introduced in this paper will focus on solving point-wise semantic segmentation problem on 3D point clouds, which will estimate a semantic label for each 3D point, as shown in Fig.\ref{fig:example}. Some previous work like \cite{zermas2017fast}, \cite{bogoslavskyi17pfg} instinctively segment point clouds based on Euclidean distance. These solutions are efficient enough, but without providing semantic information. Inspired by image-based semantic segmentation methods, some researchers　\cite{yang2018pixor},\cite{liang2018deep}, \cite{wu2018squeezeseg},\cite{wu2018squeezesegv2}, \cite{wang2018pointseg} employ mature CNN-based neural network to predict the semantic label for each pixel by projecting 3D point clouds into 2D plane. Such methods like SqueezeSeg \cite{wu2018squeezeseg} and SqueezeSegv2\cite{wu2018squeezesegv2} are real-time but not optimal because of ignoring the internal geometric information in the 3D point cloud leading to insufficient performance. Fusion-based methods \cite{qi2018frustum}, \cite{xu2018pointfusion}, \cite{ku2018joint} resolve above limitations by joining multi-information from camera and LiDAR. Nonetheless, the 2D-based detection might fail on some challenging cases that could only be well observed from 3D space and typically run slowly due to processing a significant amount of images and point clouds input. A further approach \cite{shi2018pointrcnn} is to operate the 3D data directly, using the bottom-up scheme to generate the 3D bounding box proposals and implement the standard 3D bounding boxes refinement. It achieves good results in the 3D detection task, yet it uses a deep neural network to deal with whole points in the scene, theoretically time-consuming and faces issues coordinate bias brings.

\begin{figure*}
	\begin{center}
	\scalebox{1}{
		\includegraphics[width=0.97\textwidth]{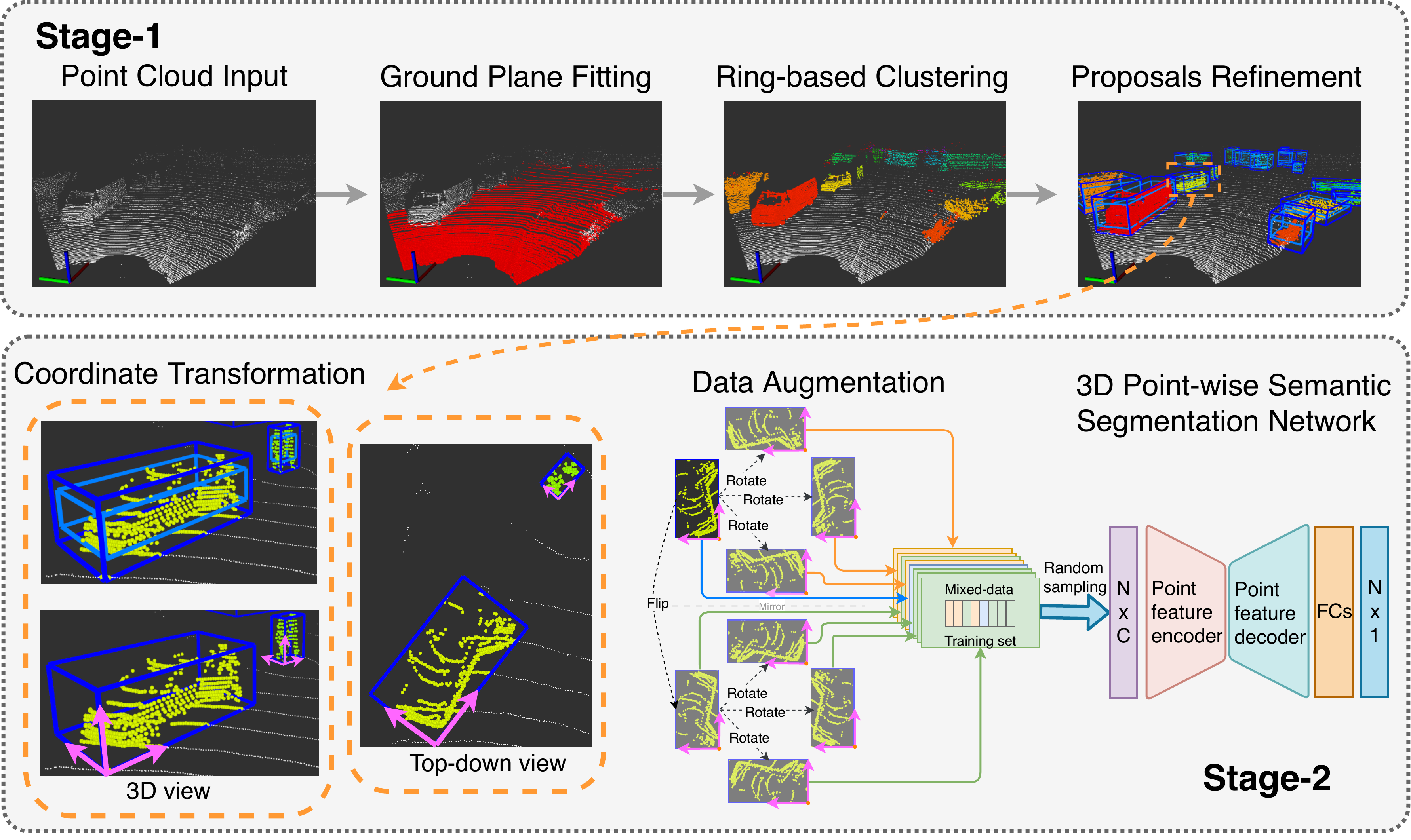}
		}
	\end{center}
	\caption{The PASS3D architecture for point-wise semantic segmentation. The whole framework consists of two stages. At stage-1, we remove the ground points, cluster point clouds without ground into several parts and then refine the proposals. At stage-2, proposals are canonical transformed and then augmented. Blue, orange and green arrows represent real, rotated and flipped data, respectively. We put them into a neural network to get the semantic results for each point.}
	\label{fig:framework}
	\vspace{-4mm}
\end{figure*}

To address the above challenges, we propose a novel two-stage framework, \textbf{PASS3D} (\textbf{P}recise and \textbf{A}ccelerated \textbf{S}emantic \textbf{S}egmentation for \textbf{3D} Point Cloud), utilizing 3D geometric clustering algorithm and 3D deep learning scheme. The framework combines the efficiency of traditional geometric methods with the robustness of advanced deep learning networks. At stage-1, we remove the ground points taking advantage of the geometric and topological structure in 3D space and segment the remains into several clusters expeditiously by a ring-based method\cite{zermas2017fast}. Then we refine the clusters to get final proposals. At stage-2, we apply the canonical transformation on the proposals with augmentation by a novel method introduced by us to eliminate coordinate bias and then put them into a powerful point set processor like PointNet++\cite{qi2017pointnet} to obtain point-wise semantic information. The overall pipeline is shown in Fig.\ref{fig:framework}.

Compared with the state-of-the-art\cite{wu2018squeezesegv2}, our method is $\textbf{16.5\%}$ better on 3D `Pedestrian' IoU, boosts 3D `Cyclist' IoU by $\textbf{17.2\%}$ and our average IoU increases by $\textbf{7.9\%}$. Our cluster proposal method at stage-1 achieves $\textbf{89.5\%}$ point-wise recall in \textbf{5ms} with only about \textbf{30} proposed clusters per frame.

The main contributions of our work are as follows:

(1)	We propose a flexible two-stage 3D semantic segmentation framework, which combines the efficiency of traditional geometric methods with the robustness of deep learning methods, obtaining pure 3D features without information loss.

(2)	Our accelerated cluster proposal algorithm achieves higher point-wise recall with less redundant proposals in an extremely short time, which dramatically shortens the whole time and reduces subsequent calculations, making it suitable for autonomous driving applications.

(3)	We introduce a novel data augmentation method for point cloud learning problem that alleviates coordinate bias in 3D space and increases the performance and generalization ability of the network, especially for non-rigid objects in non-Euclidean metric space. 

(4) Experiments on the KITTI raw dataset\cite{Geiger2013IJRR} show that our method outperforms the state-of-the-art methods with remarkable margins. Our source code will be open-sourced.

\section{Related Work}

Some previous works segment point clouds based on Euclidean distance. \cite{douillard2011segmentation} concluded several methods for removing ground points based on iterative algorithms, like RANSAC and GP-INSAC.\cite{bogoslavskyi17pfg}, \cite{moosmann2009segmentation} use a range image to compute local convexities of the points in the cloud. \cite{bogoslavskyi17pfg} proposed an effective ground segmentation and clustering algorithm, while \cite{shin2017real} directly extracted foreground objects without ground segmentation. \cite{wang2012could} focuses on the entire process, including segmentation, clustering, and classification. \cite{zermas2017fast} proposes a ring-based method, segmenting point cloud with ground removed, which is efficient for 3D LiDAR point cloud. However, the above methods are not able to provide semantic information. 

Existing works on 3D object detection or semantic segmentation based on point cloud data can be divided into three ways:

\smallskip
\noindent
\textbf{1) 2D-based methods:} Inspired by mature image-based semantic segmentation frameworks, several methods project the point cloud into the BEV (birds-eye-view) (\cite{ku2018joint}, \cite{yang2018pixor}, \cite{liang2018deep}, \cite{chen2017multi}) or FV (front-view) (\cite{wu2018squeezeseg}, \cite{wu2018squeezesegv2}, \cite{wang2018pointseg}) and use a 2D CNN to learn the characteristics of the point cloud for detection or semantic segmentation. In \cite{yang2018pixor}, a fast single-stage detector is designed, utilizing a specific-height-encoded BEV input. Only a small amount of data needs to be processed in this kind of methods. However, the key issue with these methods is that many points are discarded when generating the 2D maps, resulting in a large loss of information on the vertical or depth axis. The loss of information severely affects the performance of 3D feature learning. 

\smallskip
\noindent
\textbf{2) Fusion-based methods:} Fusion-based methods extract muti-information from 2D images and 3D point cloud. MV3D \cite{chen2017multi} takes birds-eye-view and front-view of LiDAR point cloud as well as an RGB image as input to obtain multi-channel features. In \cite{qi2018frustum}, a 2D detection network is used to proposal frustum point cloud and then PointNet++\cite{qi2017pointnet++} is applied to predict 3D object bounding box. These methods perform better than others generally but theoretically run slowly. In addition, 2D image-based proposal might fail in some challenging situations that could be well observed from 3D space.

\smallskip
\noindent
\textbf{3) 3D-based methods:} Most of the current methods are based on 3D, either converting them to 3D voxels or using point cloud data directly. Voxel-based methods like \cite{yan2018second}, \cite{zhou2018voxelnet} discretize point cloud data into voxels and then 3D convolution is applied. Improper selection of voxel size will affect performance, too large will lose details while too small will increase the amount of calculation a lot. \cite{qi2017pointnet}, \cite{qi2017pointnet++}, \cite{shi2018pointrcnn} learn point-wise features directly from the point cloud. The latter makes full use of 3D information avoiding former shortcomings. 

Inspired by \cite{zermas2017fast}, our work at stage-1 implements the ring-based clustering method to segment point cloud and refine the cluster proposals. Our semantic segmentation network at stage-2 uses 3D point cloud data directly, attaching point-wise semantic estimation. 

\begin{figure}[b]
	\begin{center}
	\scalebox{1}{
		\includegraphics[width=0.8\columnwidth]{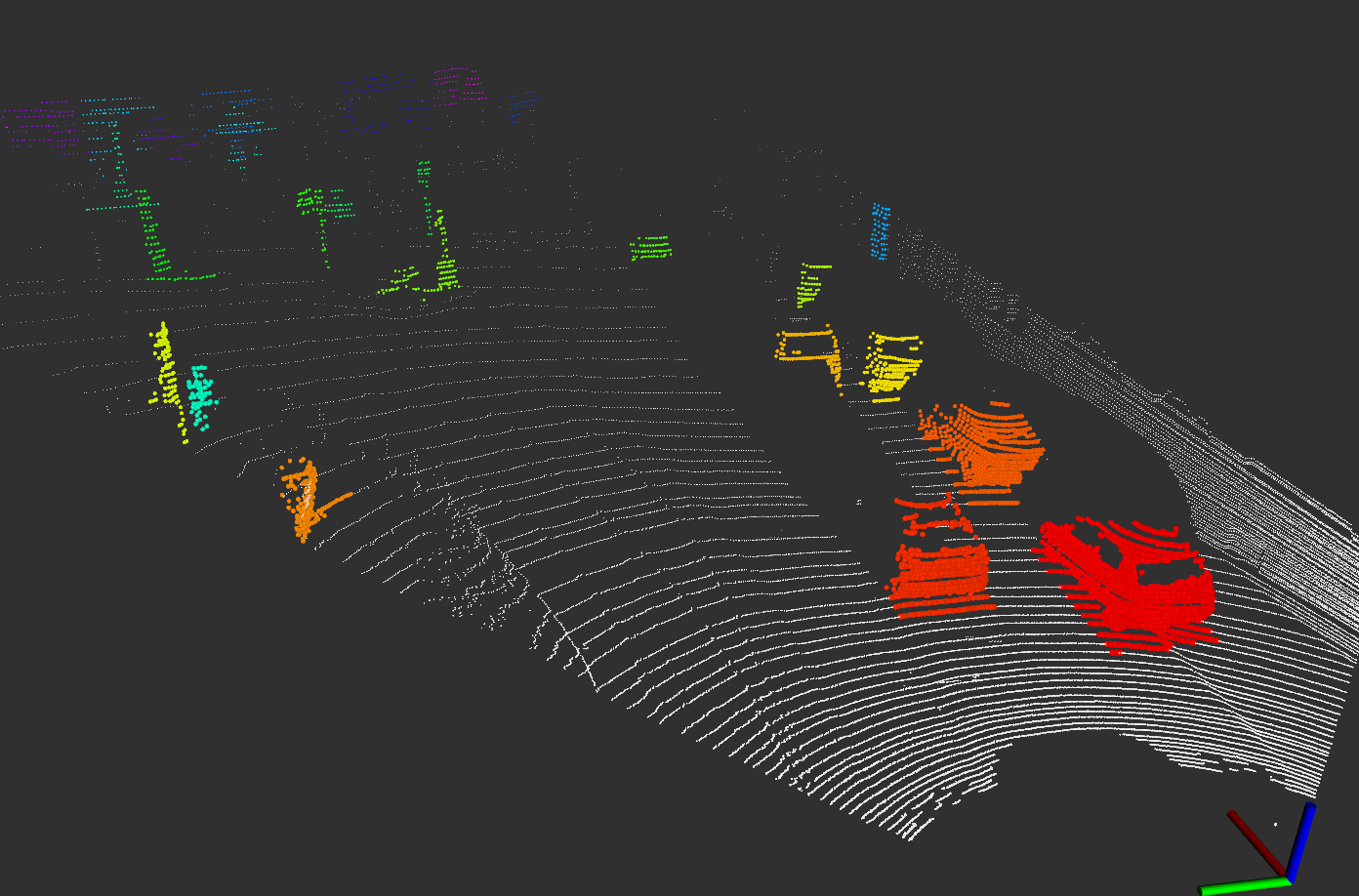}}
	\end{center}
	\vspace{-1mm}
	\caption{An instance of stage-1 results. Different cluster proposals are marked with different colours. The white points represent background, including ground and proposals filtered out.}
	\label{fig:cluster}
	\vspace{-2mm}
\end{figure}

\section{PASS3D Framework}
In this section, we present our two-stage point-wise semantic segmentation framework. The completed pipeline is illustrated in Fig.\ref{fig:framework}, which consists of the accelerated cluster proposal stage and the point-wise semantic prediction stage. Our key insight is to segment the entire scene and cluster the point clouds into multiple meaningful sub-parts and then put them into a powerful point set processor to get the point-wise classification label. We find that the objects we are interested in (such as `Car', `Pedestrian', `Cyclist') are independent in 3D space without overlap, so it is not necessary to take all the point clouds in the scene into consideration for the recognition of each object but only needs a prior, point cloud of the object itself. The point clouds after removing the ground are naturally disconnected from each other. Therefore, we think that it is feasible and efficient to obtain candidate clusters by clustering the point clouds without ground. After that, put clusters into a powerful neural network for feature extraction and semantic segmentation.

\subsection{Stage-1: Accelerated cluster proposal}

Region proposal method based on deep learning has achieved remarkable results in 2D images but performs poorly in 3D point cloud scene due to the huge 3D search space and the irregular format of point clouds. Existing methods \cite{shi2018pointrcnn}, \cite{yan2018second} promote the development of the recognition of 3D point cloud conspicuously. However, some problems still need to be solved, such as proposing too many candidates and putting all points into the neural network, which leads to a huge increase in calculations and time consumption.

We observe that objects in 3D scenes are naturally separated without overlapping from each other. For this reason, we propose an accelerated cluster proposal method to generate clusters and refine the results to get final proposals referring to \cite{zermas2017fast}, which achieves a high point-wise recall rate with few candidates in a very short time. Fig.\ref{fig:cluster} shows a piece of our cluster proposal results. This part is composed of three steps generally: ground plane fitting, ring-based clustering and proposals refinement. 

\begin{figure}
	\begin{center}
	\scalebox{1.0}{
		\includegraphics[width=0.96\columnwidth]{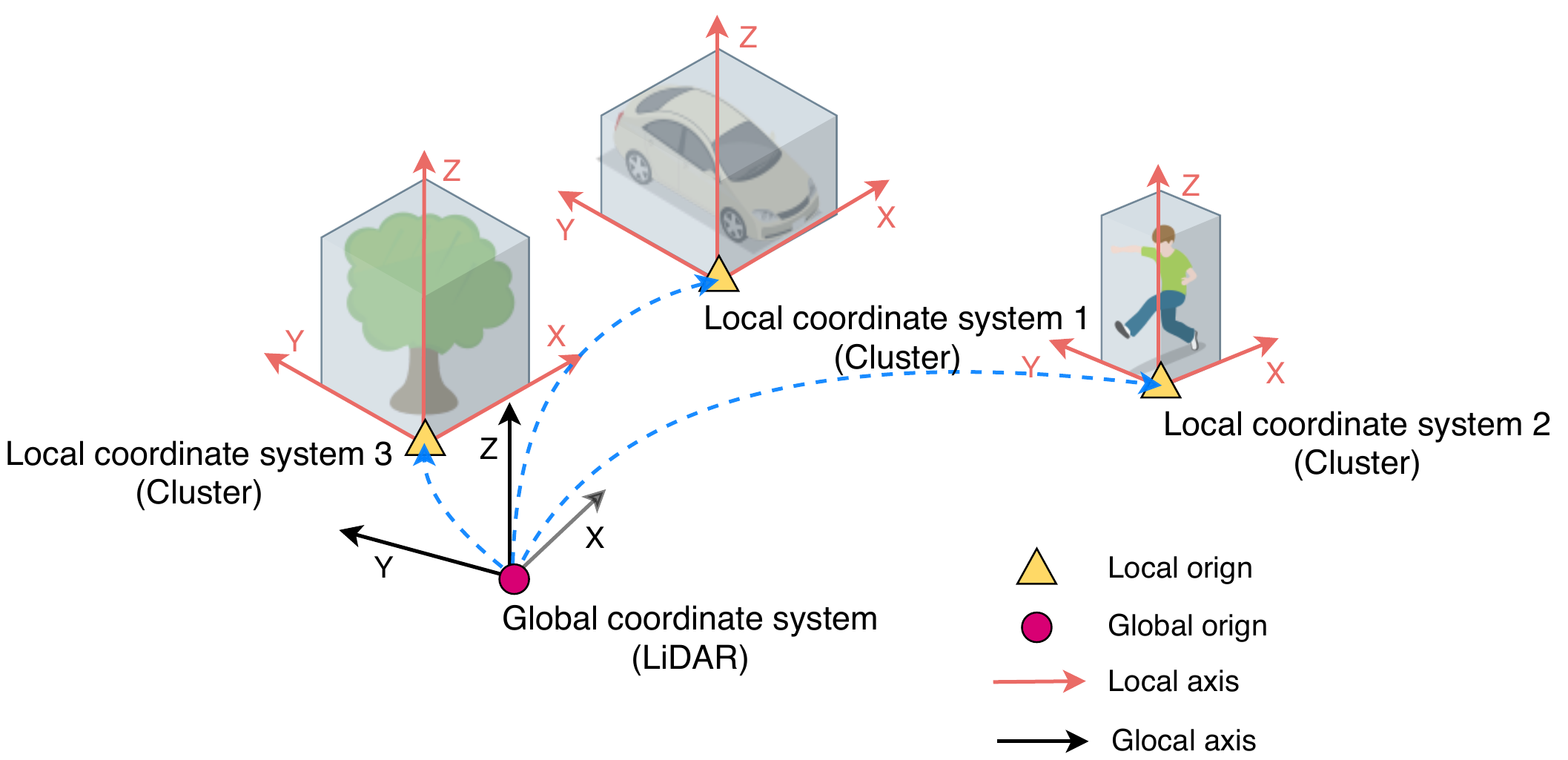}
		}
	\end{center}
	\vspace{-2mm}
	\caption{Illustration of coordinates transformation. We convert clusters' coordinates to their local systems to deal with the trouble caused by global positions' changes.}
	\label{fig:transform}
	\vspace{-2mm}
\end{figure}

\smallskip
\noindent
\textbf{1) Ground plane fitting:} Ground points removal can separate objects from each other in the 3D Euclidean space and significantly reduce the number of points involved in the proceeding computations. The distribution of points belonging to the ground is very regular and easily identifiable in the autonomous driving scene, on account of some prior knowledge:

\noindent
(i) They can be described by planes with a simple mathematical model.
\noindent
(ii) It can be acceptably assumed that points with the lowest height values are most likely to belong to the ground surface.

In reality, it is insufficient to fit the ground with one single plane due to ground fluctuations and measurement noise, so we first divide the whole scene into \textit{$N_{seg}$} segments along the moving direction of the LiDAR. Then extract a set of seed points with low height values and estimate the initial plane model of the ground for each segment. Each point will be judged whether it belongs to the estimated plane model by calculating the vertical distance between them and comparing the distance to a threshold \textit{$Th_{dist}$}. The points belonging to the ground surface are used as new seeds for refining the plane model, which repeats for \textit{$N_{iter}$} number of times. At last, we can obtain the entire ground plane by concatenating each segment derived from this algorithm. We will pass the point cloud with ground removed to the next step.

\smallskip
\noindent
\textbf{2) Ring-based clustering:} KITTI raw data\cite{Geiger2013IJRR} are obtained by Velodyne HDL-64E LiDAR, have 64 rings and are recorded in an ordered manner. The points of one ring follow the points of another in the direction of laser rotation. Thus rings can be separated by tracing the change of the quadrant. After getting the ring to which each point belongs, we can use the ring-based method \cite{zermas2017fast} to cluster the points. Points in the same ring are clustered together if their distance is smaller than the threshold \textit{$Th_{ring}$}. The label will be propagated to the point of a new ring if the distance between the point of the new ring and the point of the previous one is less than the threshold \textit{$Th_{prop}$}. When many points in the same ring have the nearest neighbours with different labels, their labels will be merged to the smallest one. On the contrary, the point will receive a new label when no appropriate nearest neighbours can be found for any of the points in the ring. Finally, we successfully divide the entire scene into several categories, and each point in the point cloud will have a clustering label. 

\smallskip
\noindent
\textbf{3) Proposals refinement:} We generate a minimal 3D oriented bounding box for each cluster and guarantee that its z-axis is perpendicular to the ground. Since we have the prior information about interested objects (e.g. `Car', `Pedestrian', `Cyclist'), we will filter the proposals according to the number of points in the cluster and bounding box size of each proposal, setting the unsatisfied clusters to the background. The adaptive threshold $Th_{num}$, number of points in clusters, descends as the distance between clusters and LiDAR increases, on account of the distribution of point clouds is sparser at the further distance.

 \vspace{-2mm}
 \begin{align}\label{Th}
 d &= \sqrt[2]{\left(x^{G}\right)^{2}+\left(y^{G}\right)^{2}+\left(z^{G}\right)^{2}}
 ,
 \\
 &\textit{$Th_{num}$} \propto \frac{1}{d}\nonumber
 \end{align}

 \noindent $\left(x^{G}, y^{G}, z^{G}\right) \in \mathbb{R}^{3}$ represents the coordinate in LiDAR coordinate system and $d$ is the Euclidean distance between point and LiDAR. We spot that when fitting the ground plane, some points belonging to objects (e.g. car wheels, human feet, the bottom of the signs) are wrongly counted as ground due to being too close to the ground surface. For this reason, we enlarge the 3D oriented bounding boxes and merge more points to the refined proposals.

\subsection{Stage-2: Point-wise semantic segmentation}
\label{ssec:pooling}

In this section, we aim at predicting the category of each point, using the refined cluster proposals from stage-1. This part is composed of two steps generally: data preparation and learning-based semantic segmentation.

\begin{figure}
	\begin{center}
		\includegraphics[width=0.96\columnwidth]{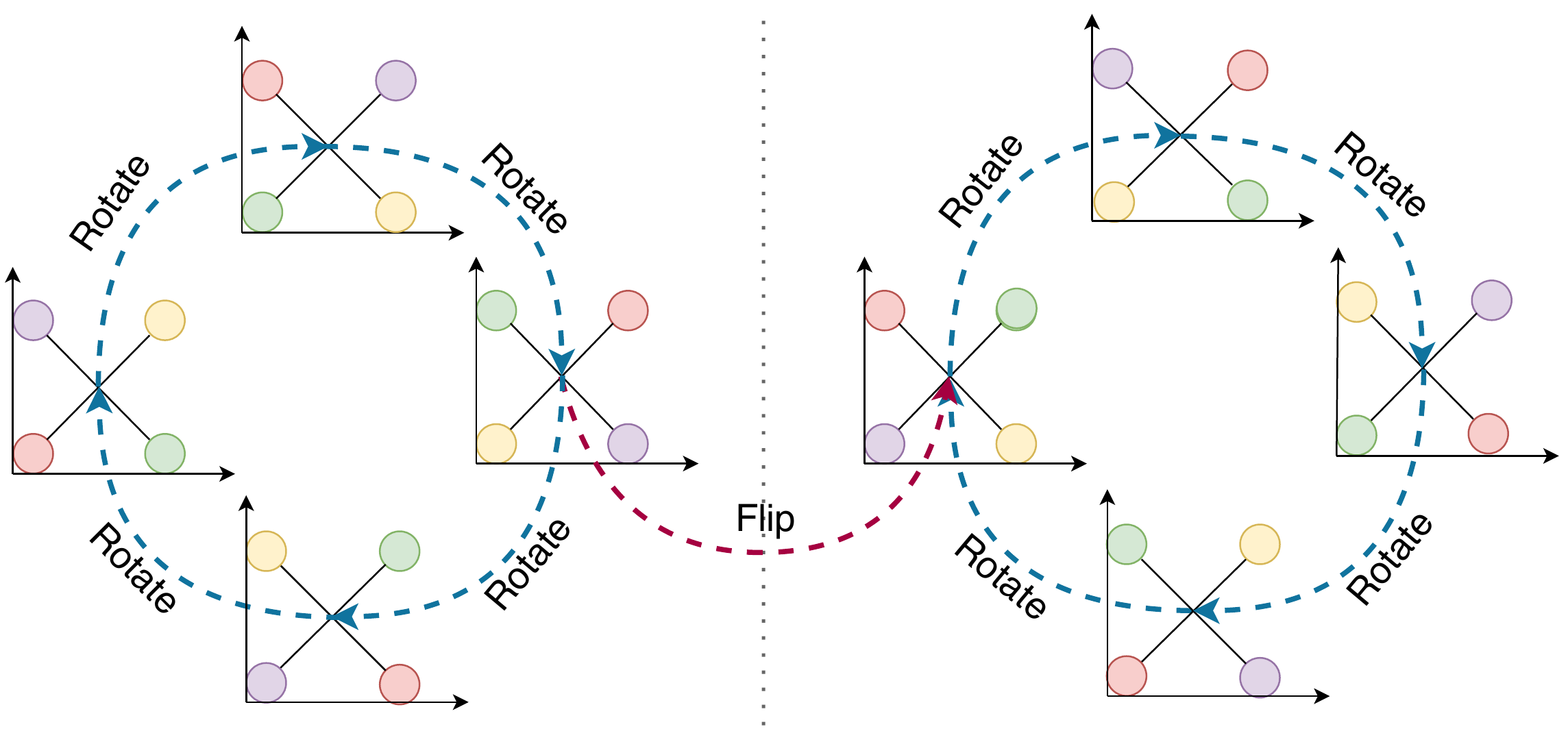}
	\end{center}
	\vspace{-2mm}
	\caption{A sketch of data augmentation. In the local coordinate system, there are eight ways to express objects in a 2D plane. Due to the asymmetry, let the four circles with different colours represent the vertexes of a bounding box. We can know that every object is unlike the others when put it under its local coordinate.}
	\label{fig:data}
	\vspace{-2mm}
\end{figure}

\smallskip
\noindent
\textbf{1) Data preparation:} 

\smallskip
\noindent
\textit{Coordinate transformation} Since the targets are distributed at various positions of the scene in the LiDAR coordinate system, the coordinates of objects change dramatically, which makes the neural network hard to converge. Considering of this, we take each cluster as a sample, randomly use one of the bottom vertexes of its 3D oriented bounding box as the origin of the local coordinate system, and put the bounding box in the first octant, which is explained by Fig.\ref{fig:transform}. The change of coordinate system will not affect the relative positions between points and make the data distribution more concentrated so that the neural network is more concerned about the relative locations rather than the absolute positions of points.

 \begin{figure*}
 	\centering
 	\small
 	\scalebox{1.47}{
 		\begin{tabular}{@{\hspace{0.0mm}}c@{\hspace{1.0mm}}c@{\hspace{1.0mm}}c@{\hspace{1.0mm}}c@{\hspace{1.0mm}}c@{\hspace{1.0mm}}}
 			
 			\includegraphics[width=0.25\columnwidth]{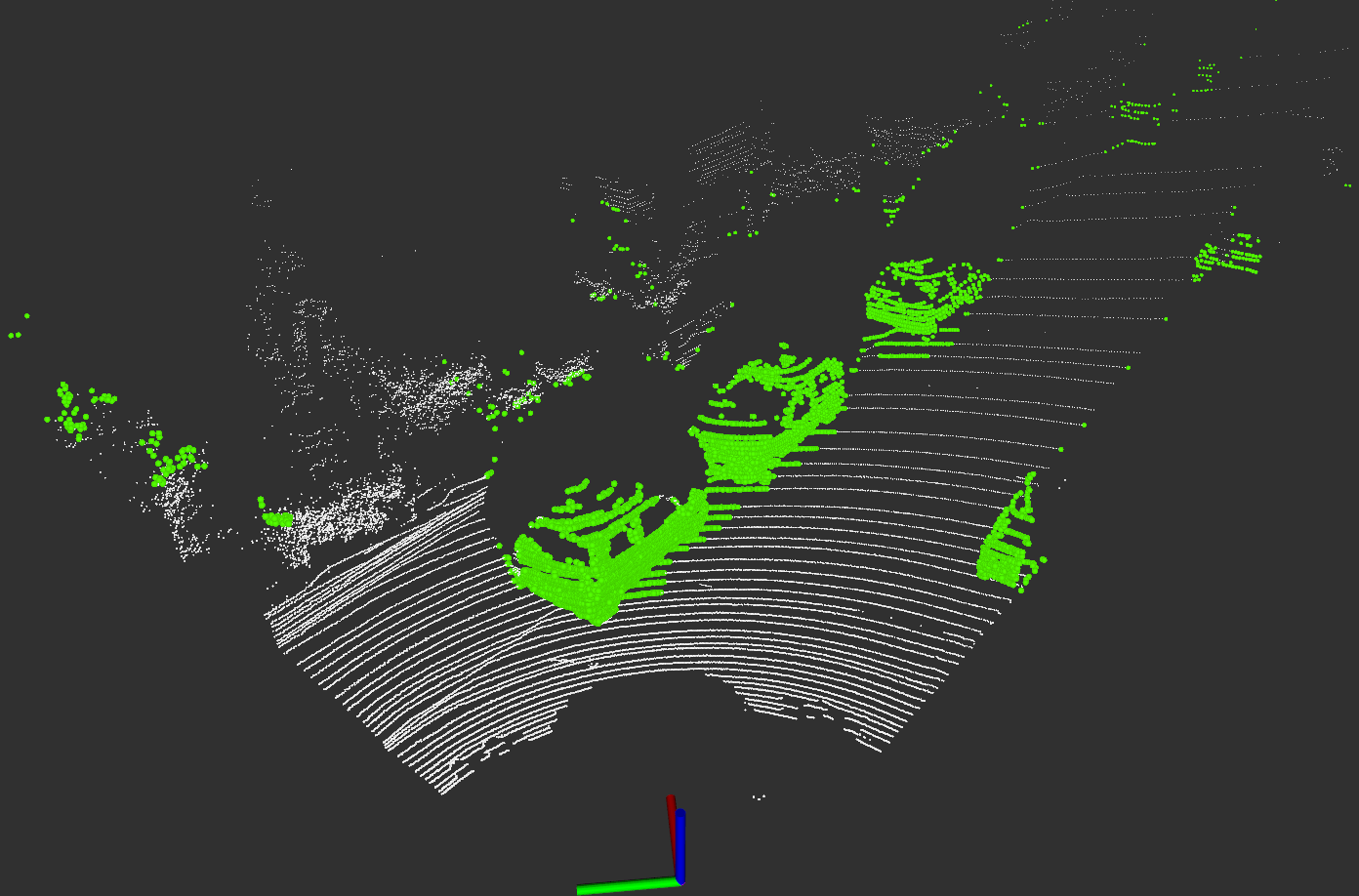}&
 			\includegraphics[width=0.25\columnwidth]{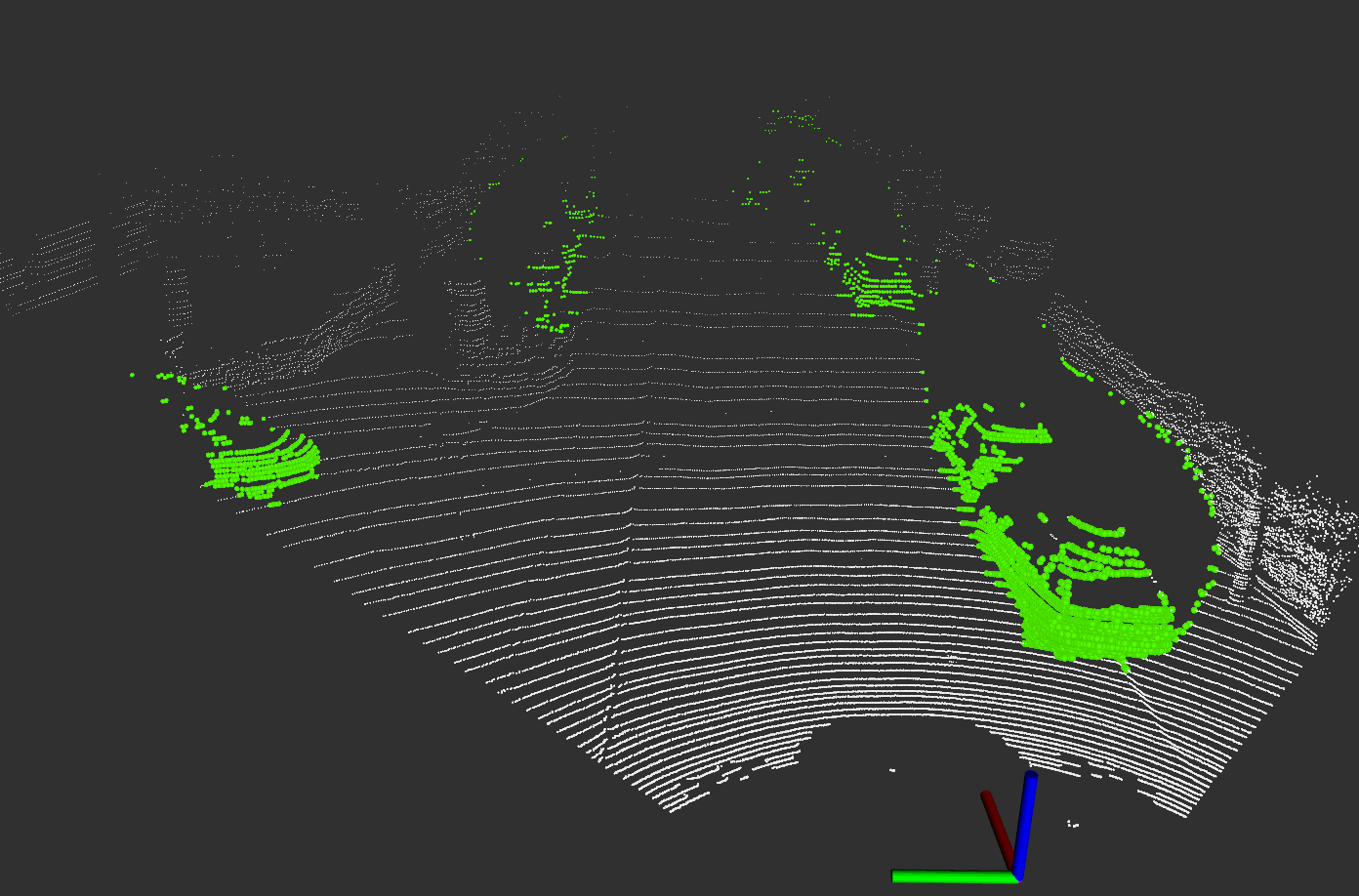}&
 			\includegraphics[width=0.25\columnwidth]{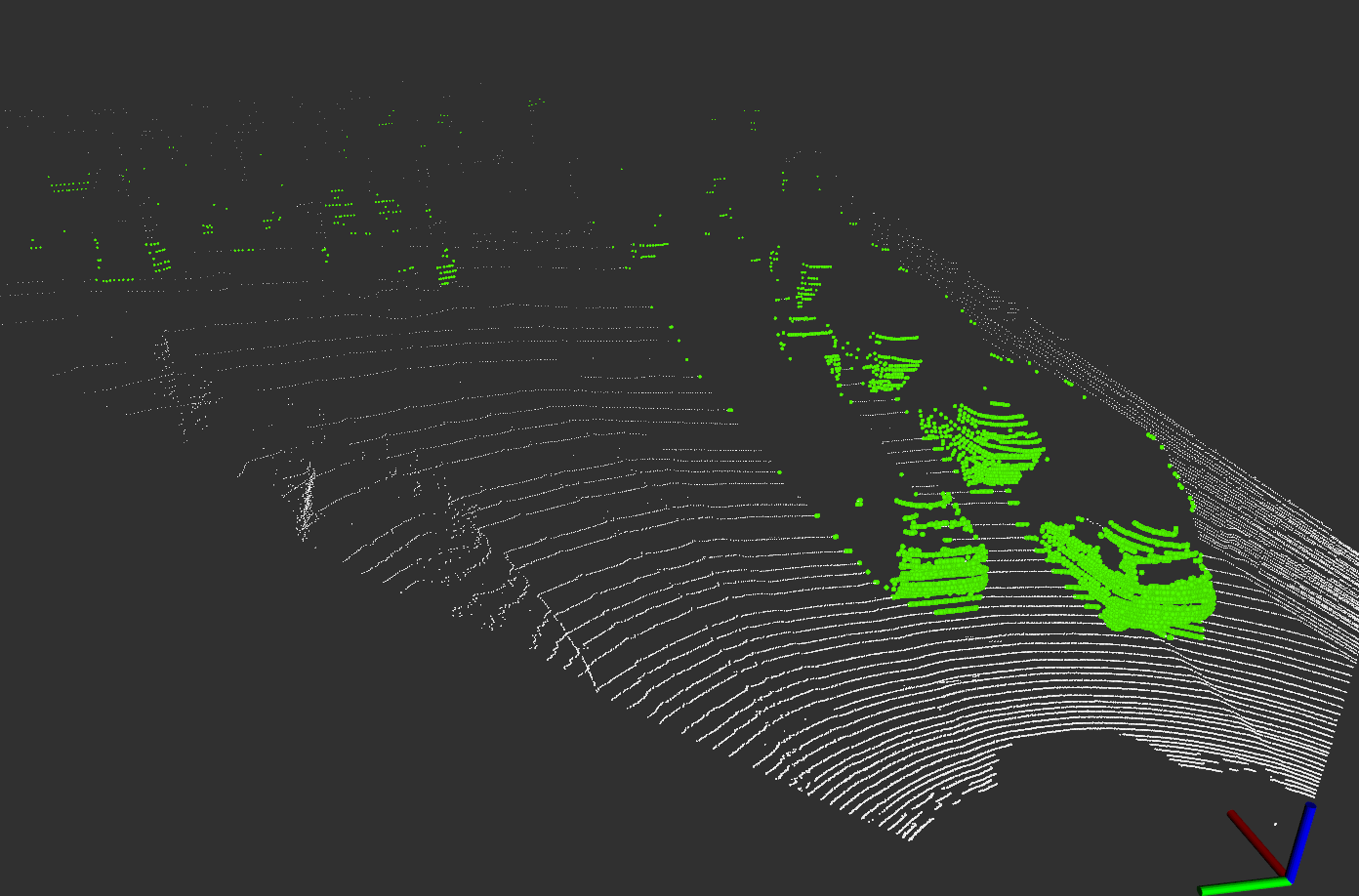}&
 			\includegraphics[width=0.25\columnwidth]{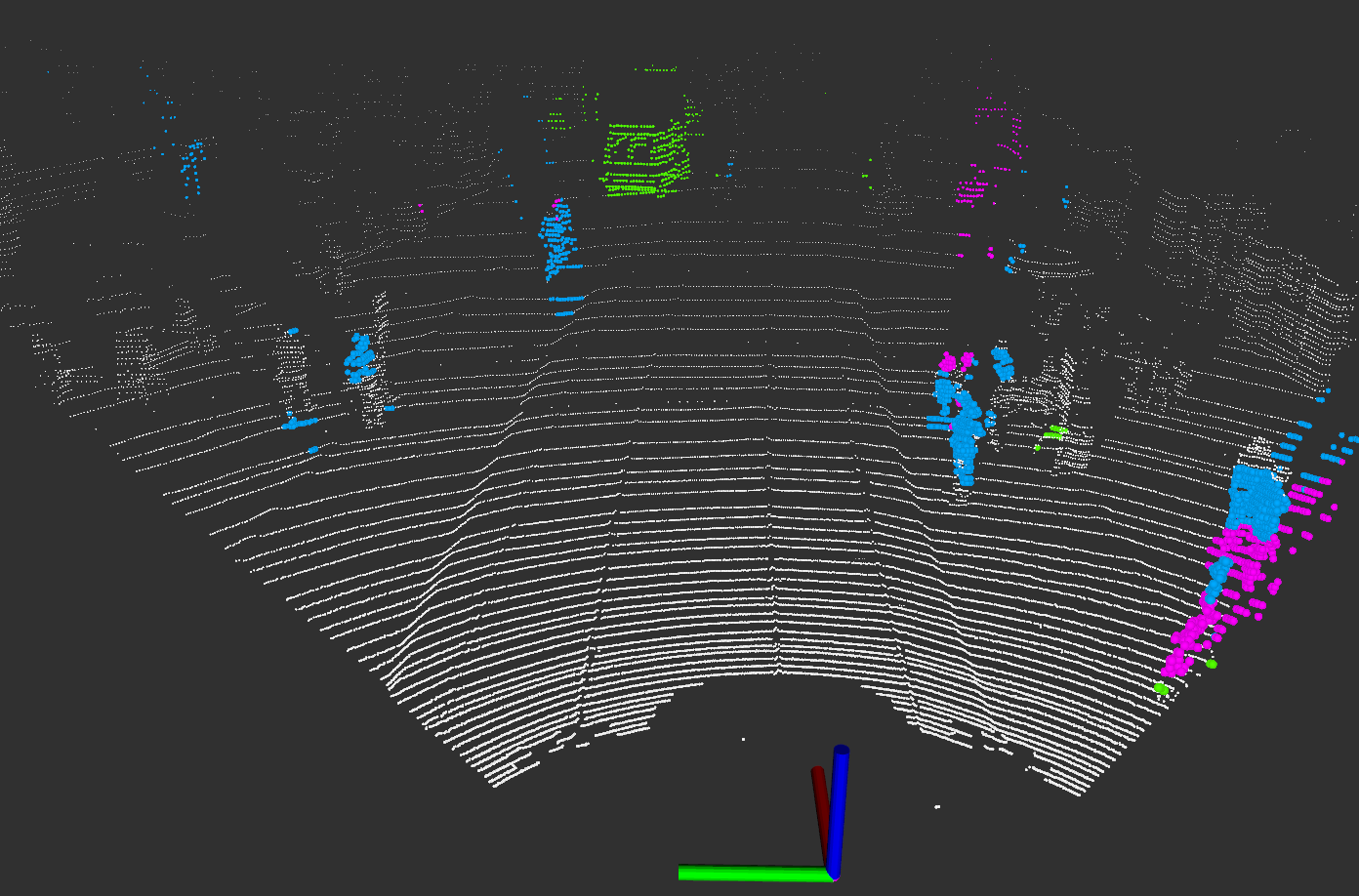}&
 			\includegraphics[width=0.25\columnwidth]{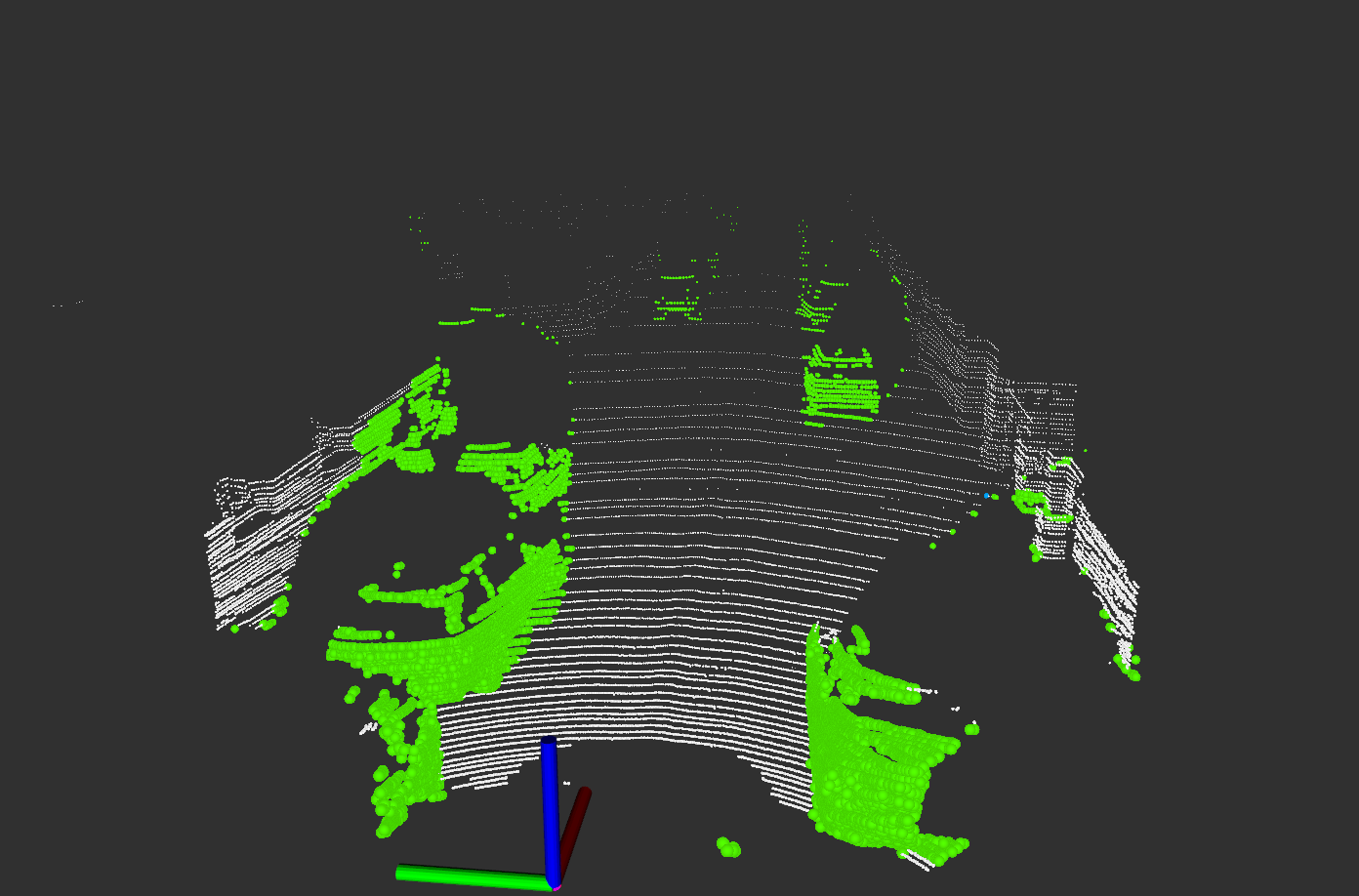}\\
 			\includegraphics[width=0.25\columnwidth]{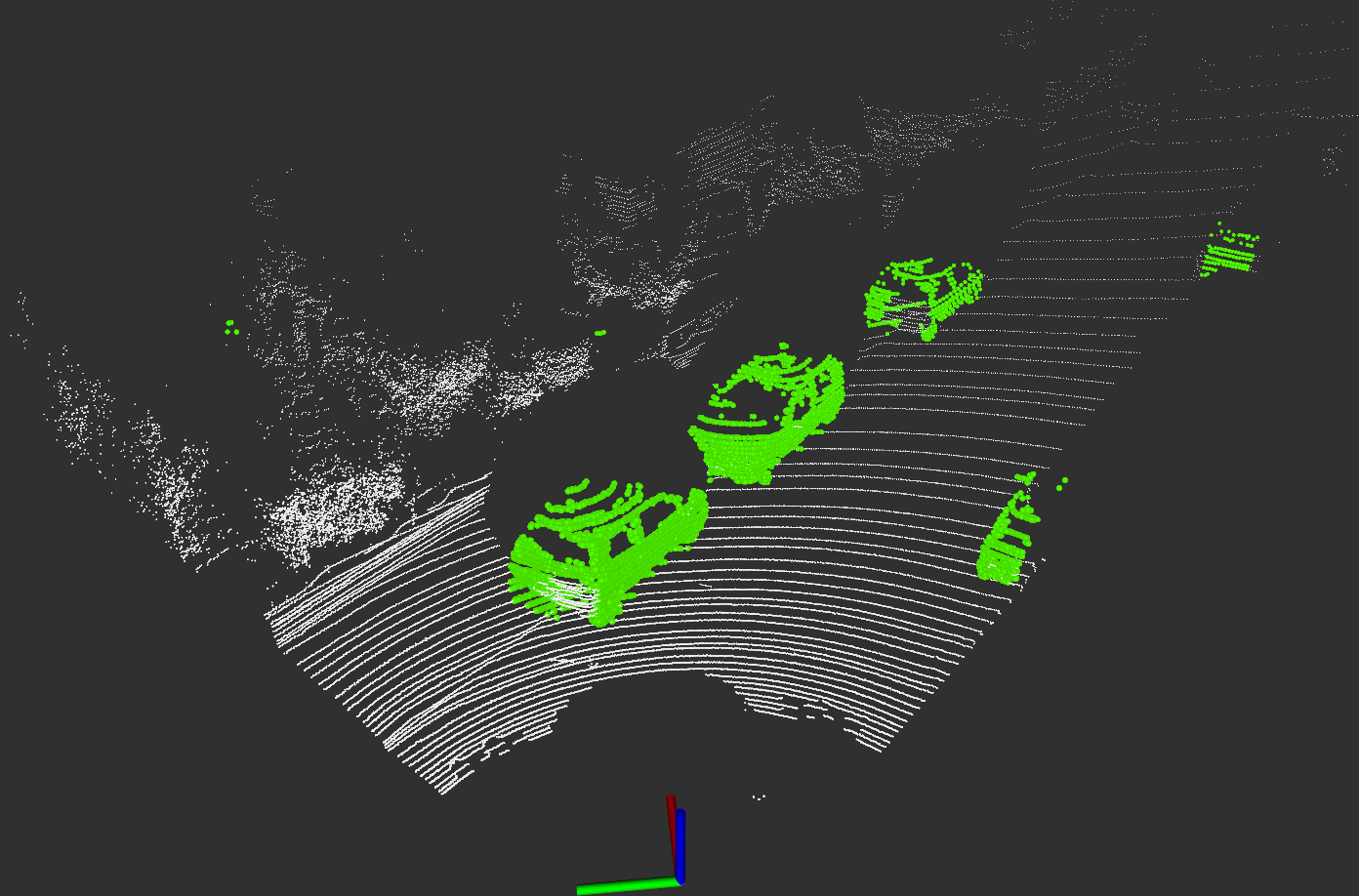}&
 			\includegraphics[width=0.25\columnwidth]{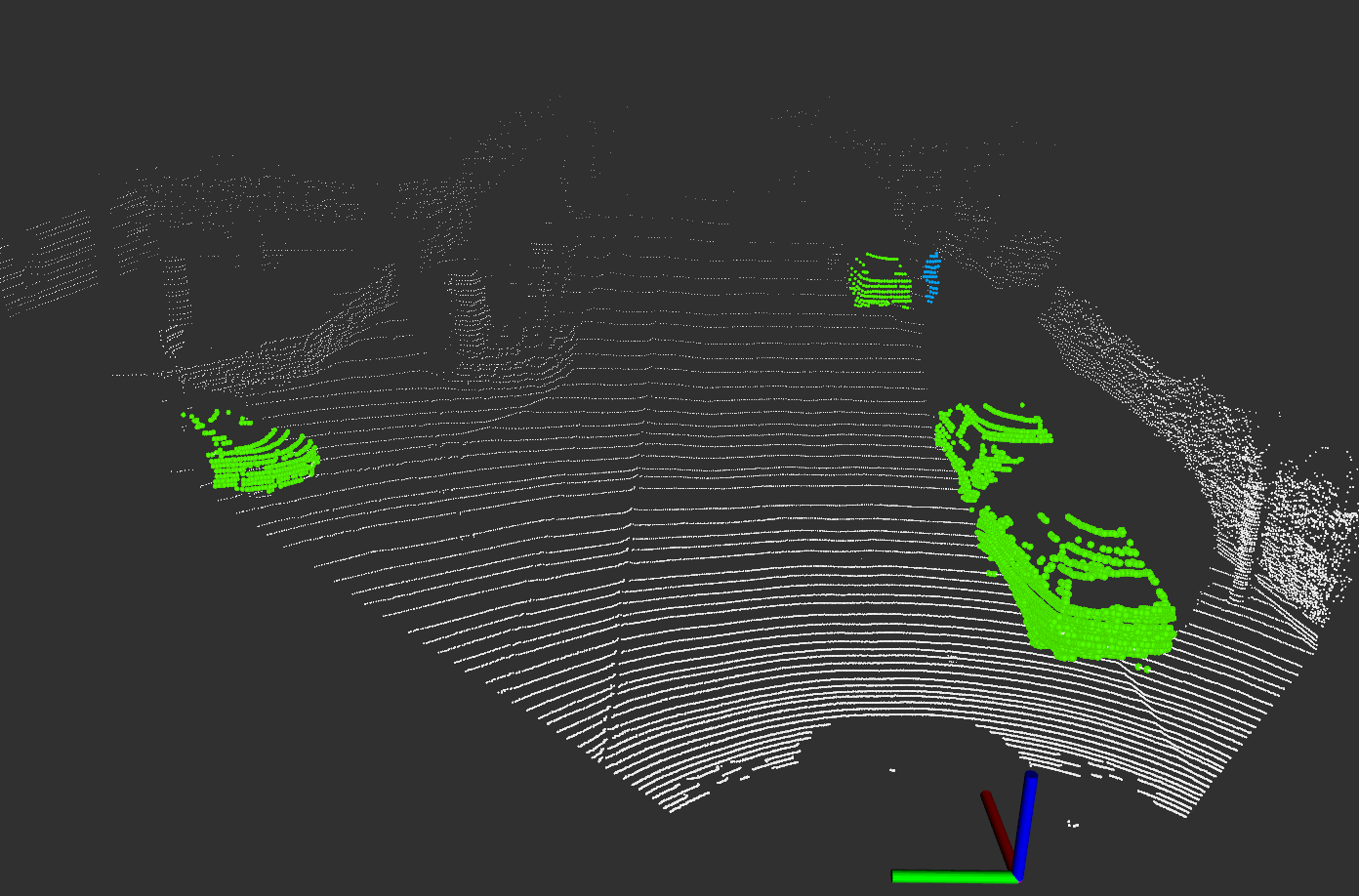}&
 			\includegraphics[width=0.25\columnwidth]{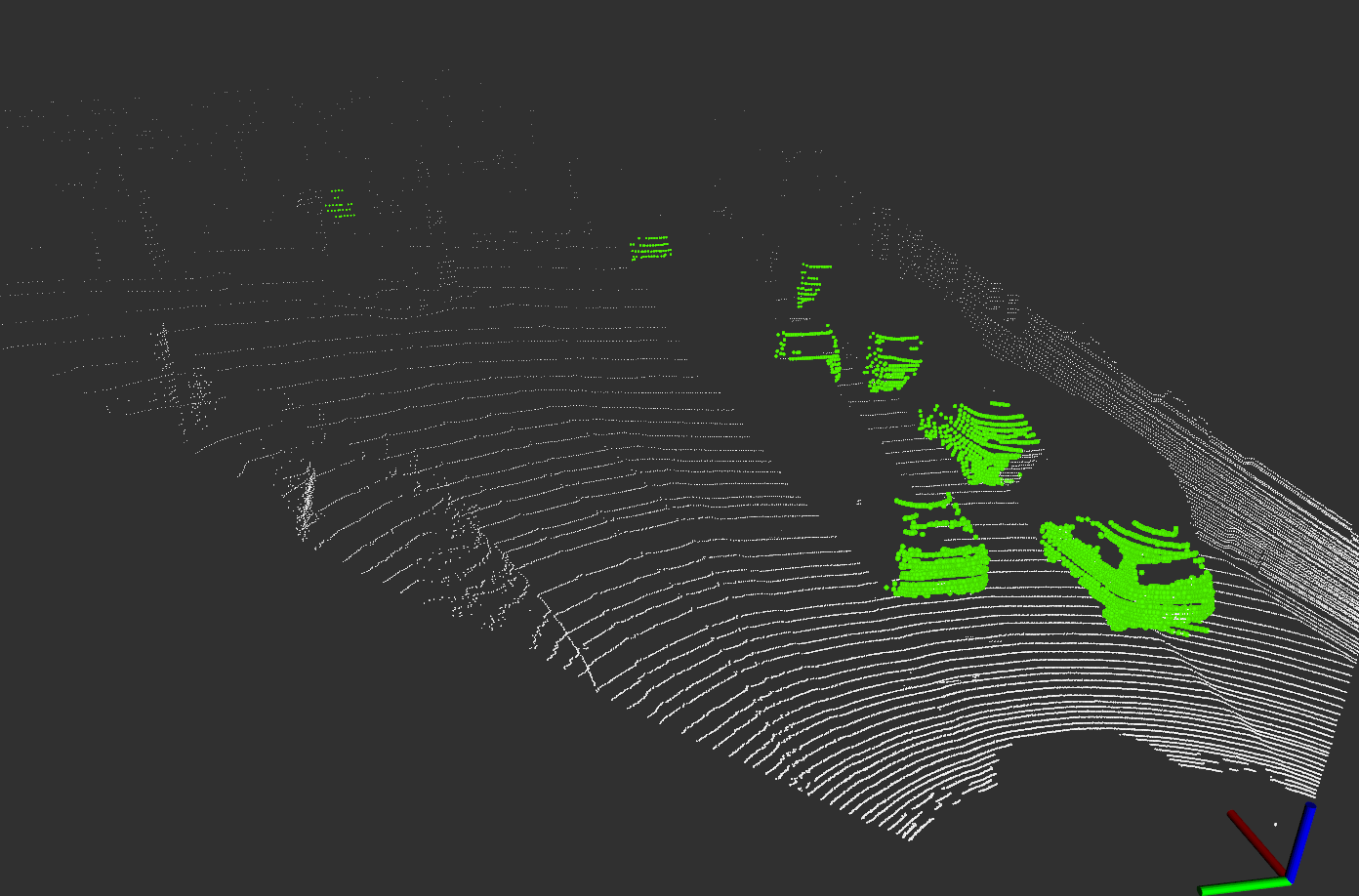}&
 			\includegraphics[width=0.25\columnwidth]{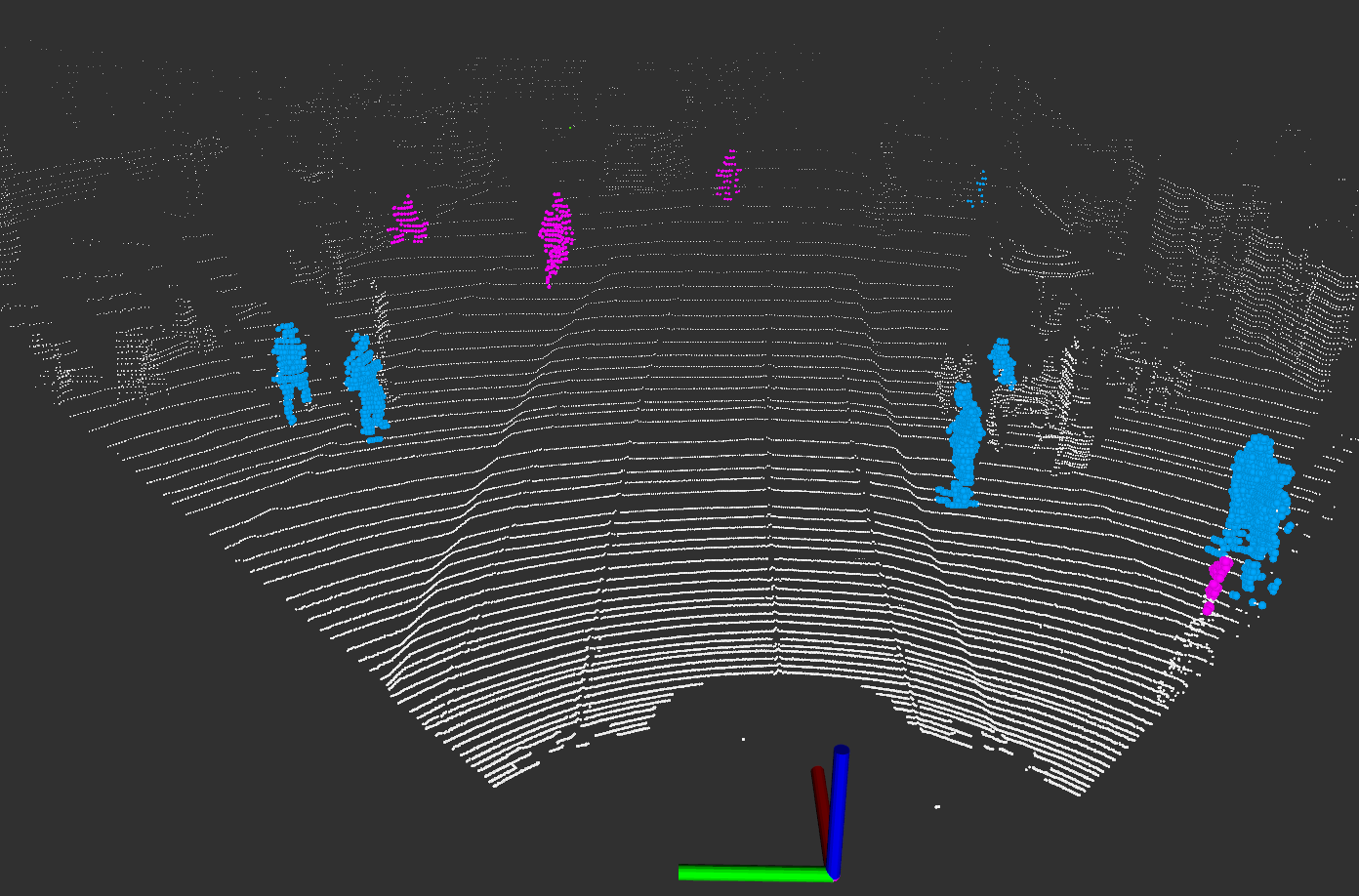}&
 			\includegraphics[width=0.25\columnwidth]{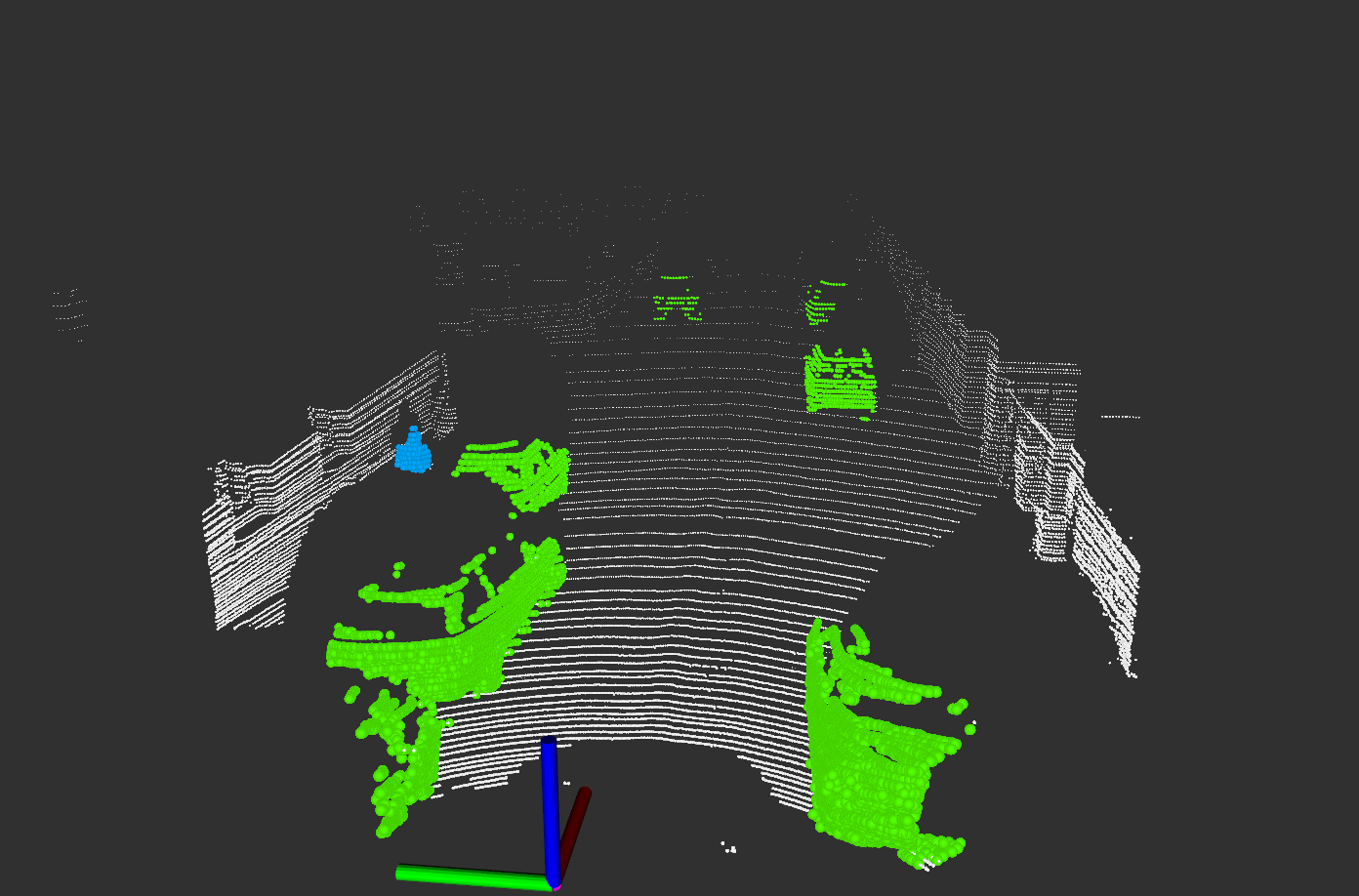}\\
 			\includegraphics[width=0.25\columnwidth]{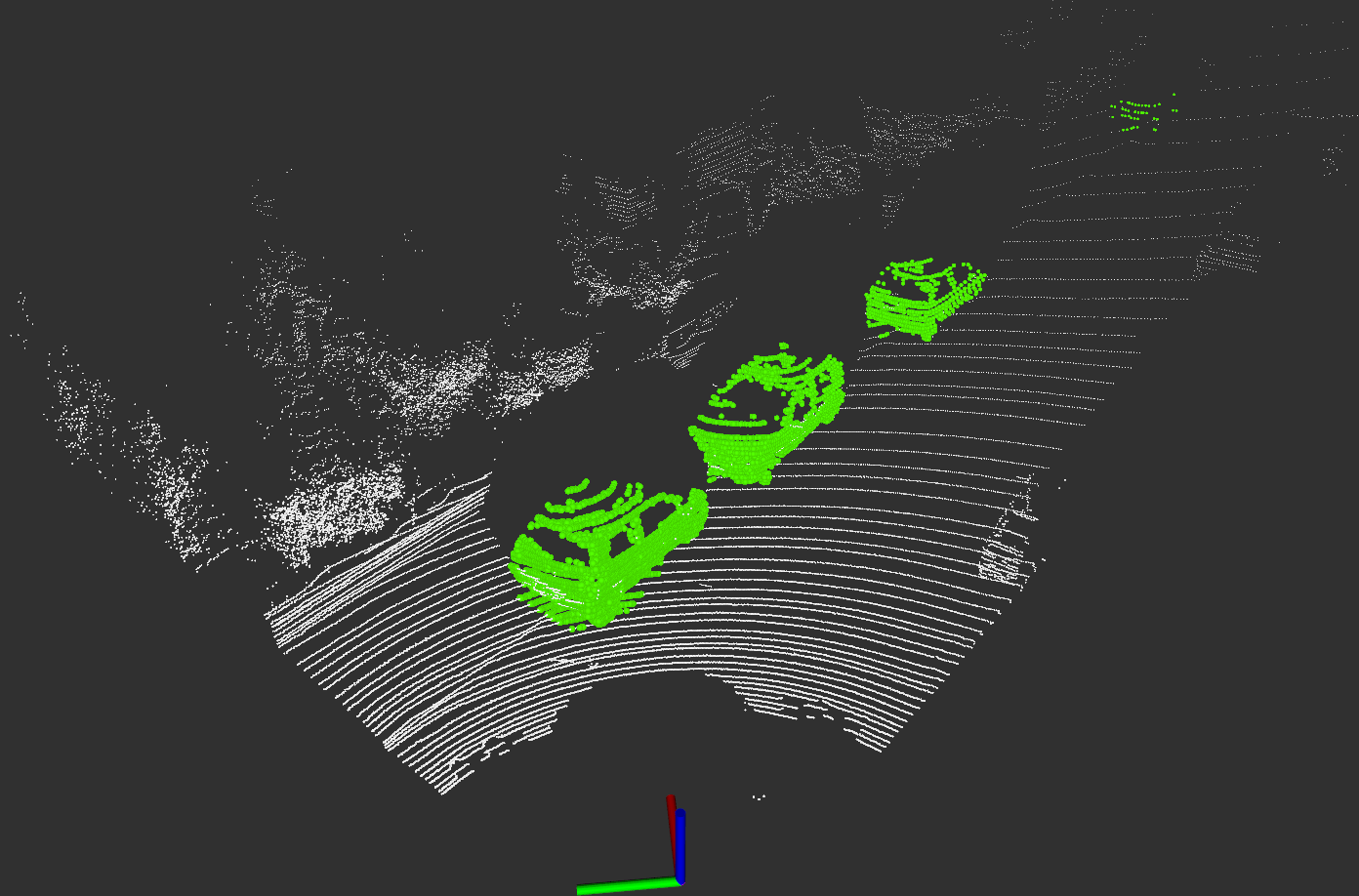}&
 			\includegraphics[width=0.25\columnwidth]{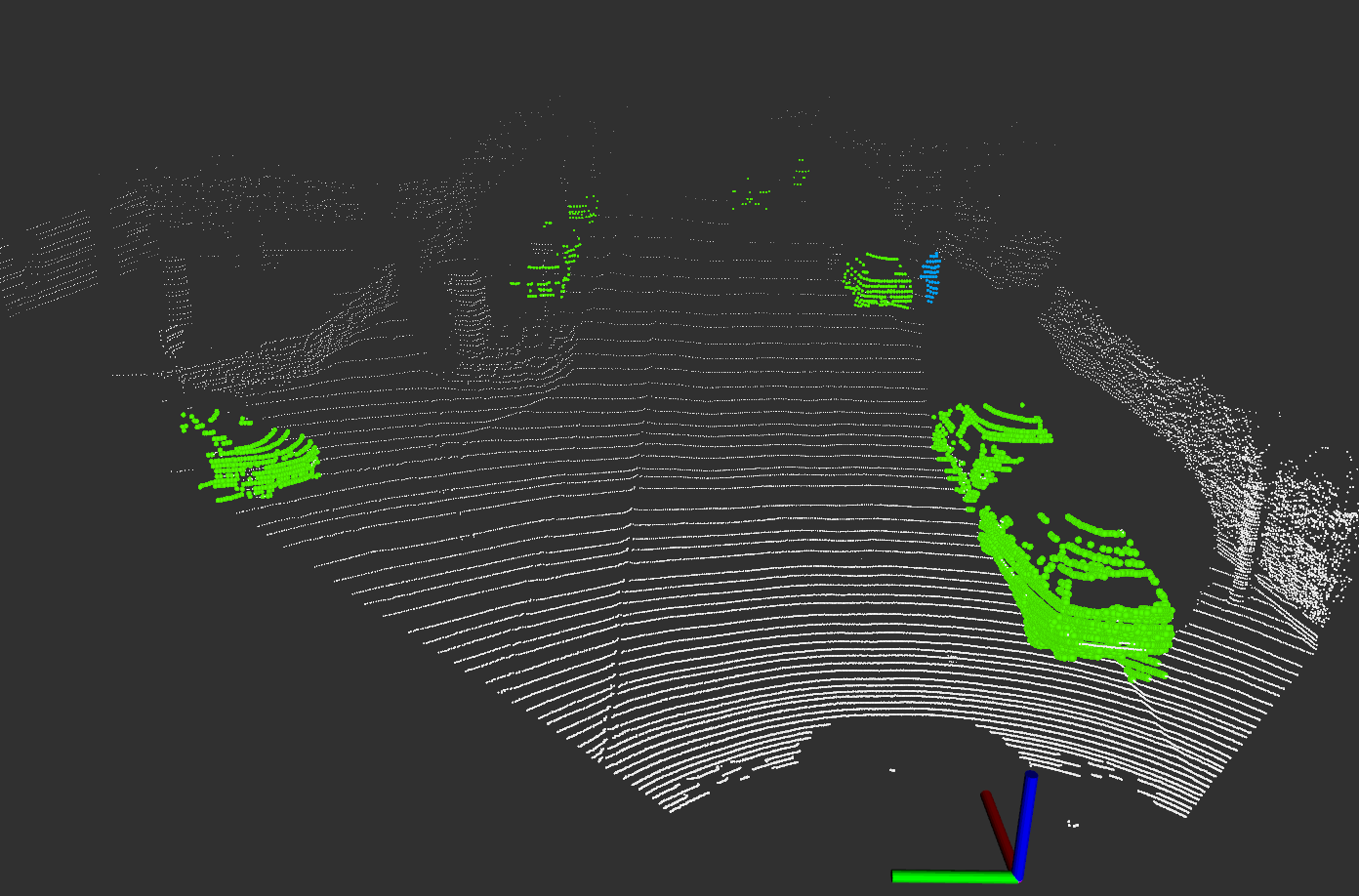}&
 			\includegraphics[width=0.25\columnwidth]{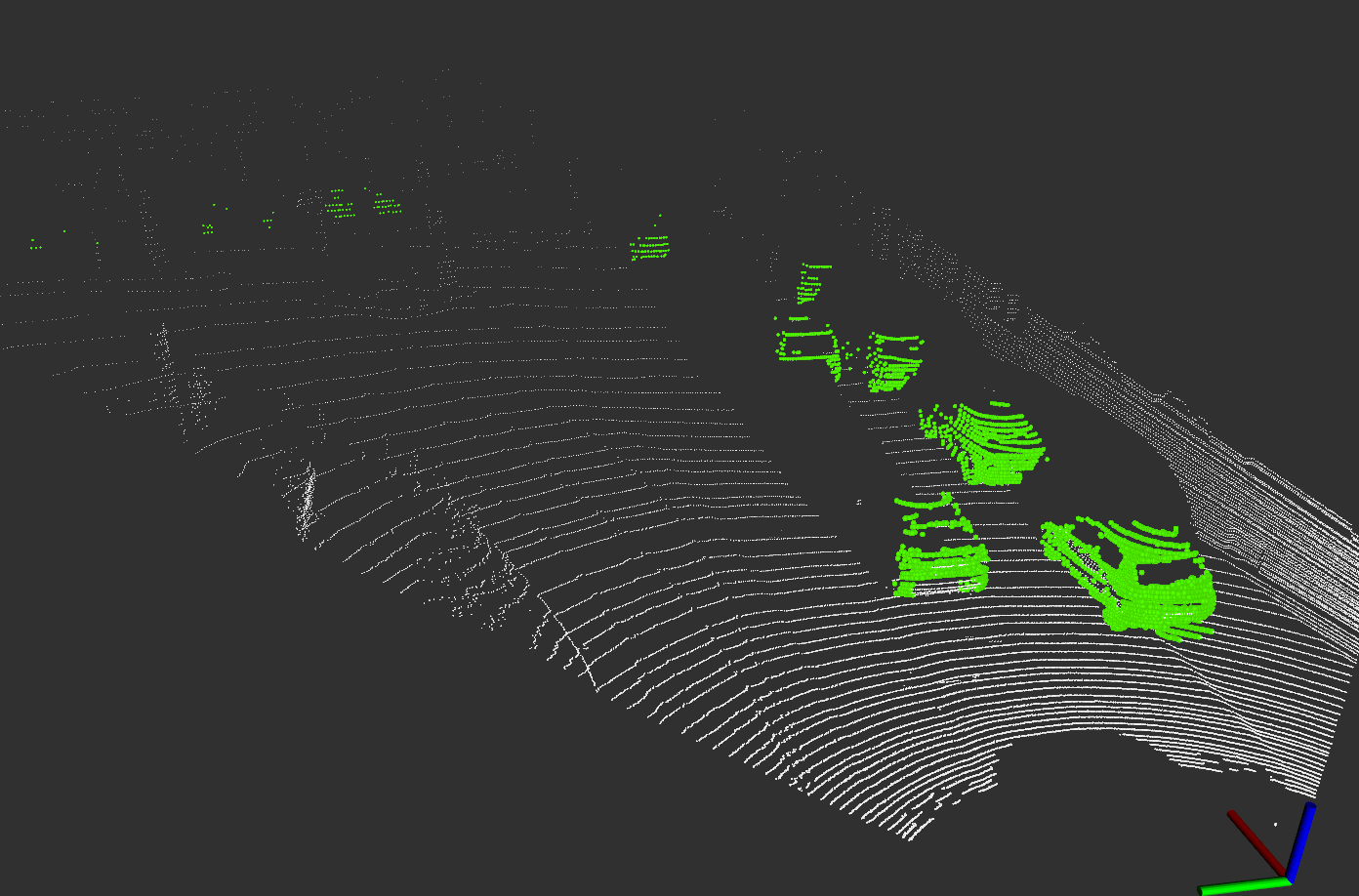}&
 			\includegraphics[width=0.25\columnwidth]{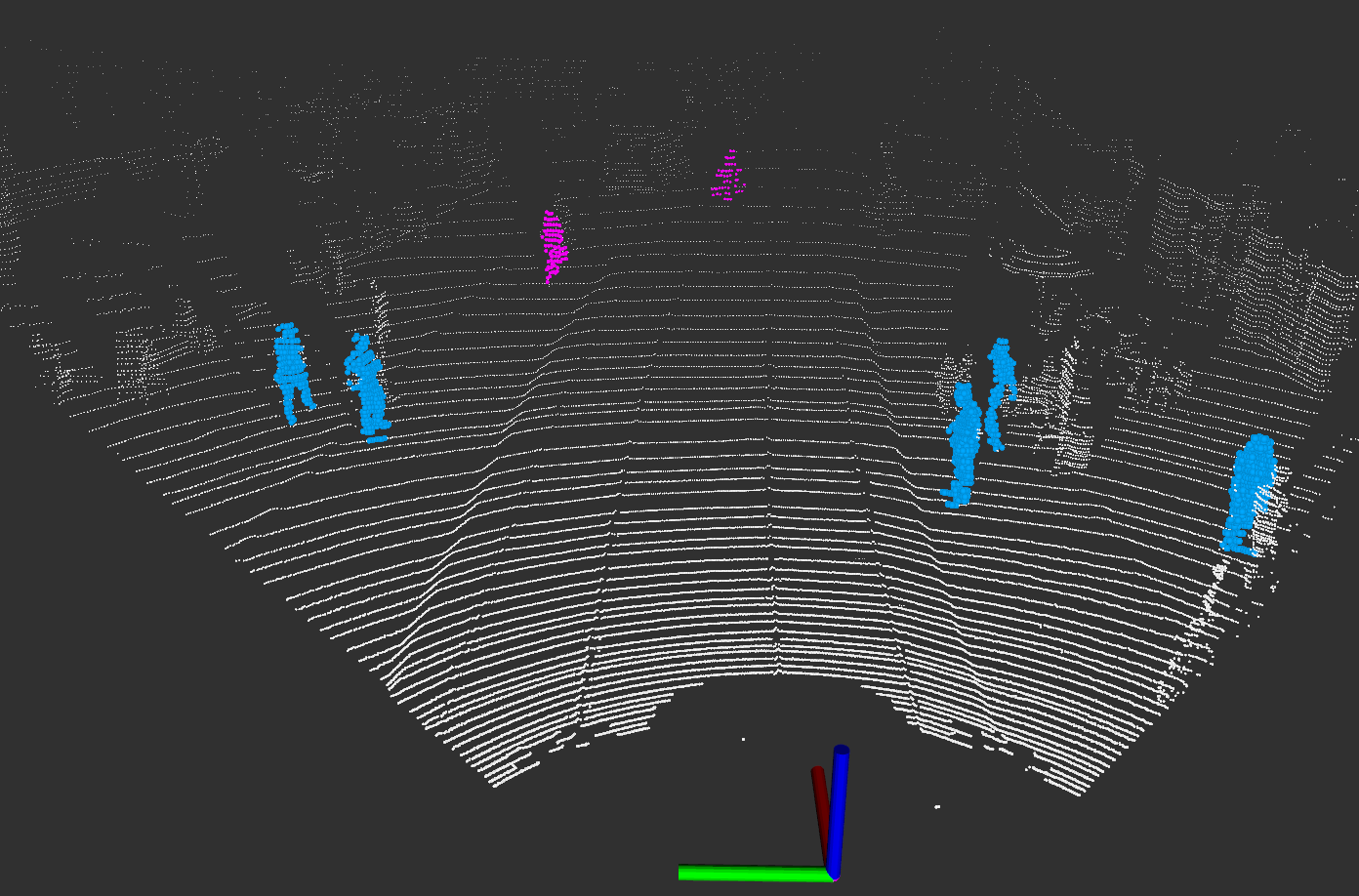}&
 			\includegraphics[width=0.25\columnwidth]{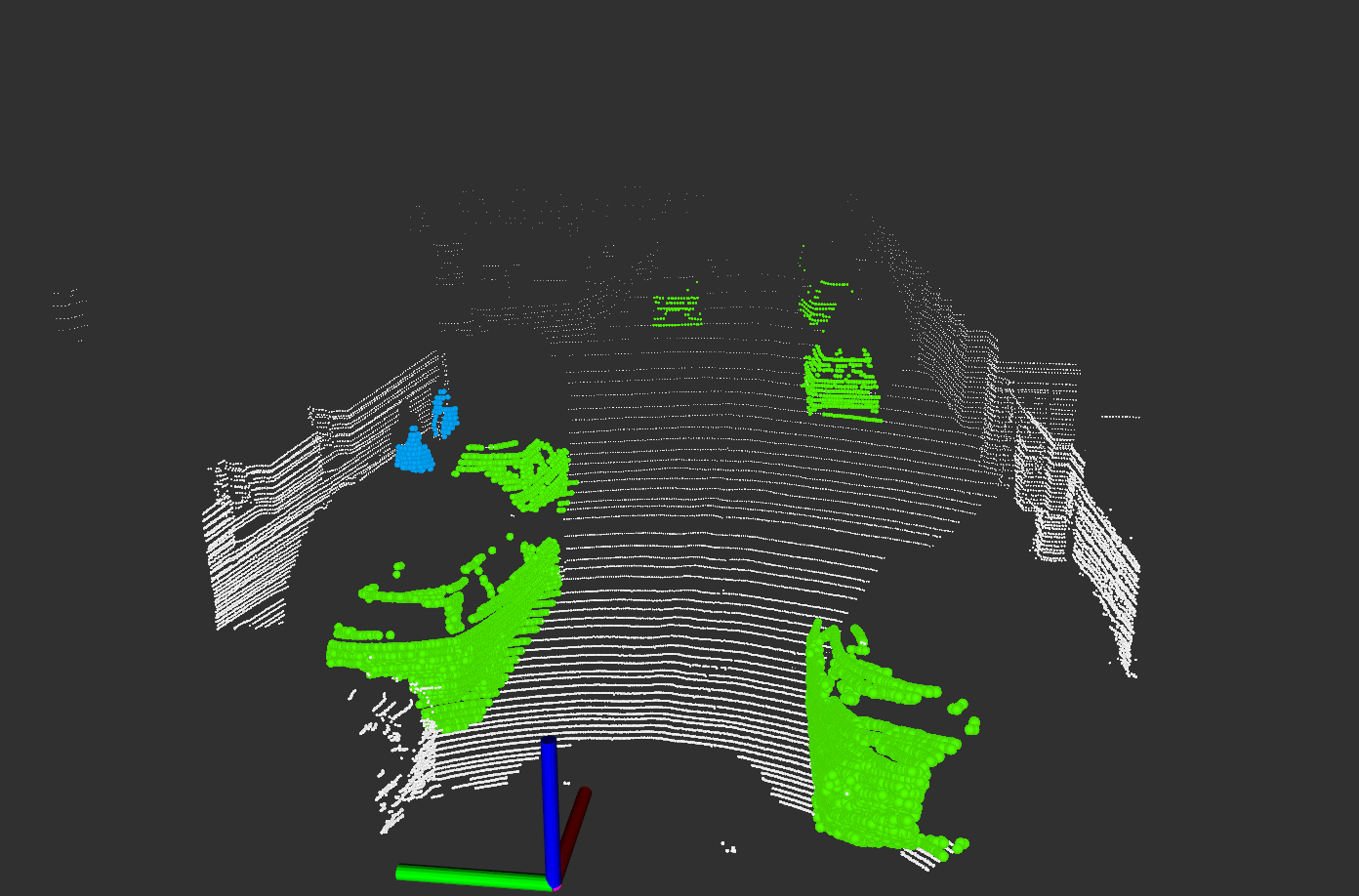}
 	\end{tabular}}
 	\caption{Qualitative results of PASS3D on our test split. The first row shows the predicted results produced by SqueezeSeg\cite{wu2018squeezeseg}, the second row is our predicted results, and the third row is the ground truth. `Car', `Cyclist' and `Pedestrian' are shown in green, pink and blue. Notice that PASS3D can additionally and accurately segment objects unlabeled in the ground truth. (Best viewed with zoom-in.)
 	}
 	\label{fig:test_vis}
 \end{figure*}

\smallskip 
\noindent
\textit{Data augmentation} Inspired by the methods of augmenting training data in some 2D or 3D CNN work \cite{zhai2019poseconvgru}, \cite{yan2018second}, we introduce a novel data augmentation method for point cloud learning problem. We discover that the imbalance of point cloud distribution in the local coordinate system will affect the generalization ability of the neural network. For example, cars in training samples with heading directions along the local x-axis are more than ones along the local y-axis, which should not affect the neural network. In order to suppress the adverse effects caused by the uneven distribution of point clouds, unlike other methods \cite{yan2018second},\cite{shi2018pointrcnn} usually augmenting the whole scene, we propose a data augmentation method dealing with our proposals, which is more efficient and targeted. As shown in Fig.\ref{fig:data}, a sample has eight representations in the local coordinate system totally (ignoring vertical direction). We rotate and mirror the sample to create the remaining seven generated samples without changing the sample category. All these eight samples are possibly obtained by our stage-1 method in the real world. We mix these eight samples into the training set of the network, which will be randomly sampled during the training process. In this way, the learning-based method could be insensitive to changes of view-point (the local coordinate system selection) and the negative influences of coordinate bias could be alleviated to some extent.
Our insight behind this is that the shape of non-rigid objects (like pedestrians and cyclists) is variable, and each of these samples is unique and rare at each moment. By employing this method, we enrich the training data, which are non-repetitive and asymmetric. Noticeable improvements to these objects are verified in our experiments.

\smallskip
\noindent
\textbf{2) Learning-based semantic segmentation:} Our network takes prepared clusters as input and predicts a probability score for each point that indicates how likely the point belongs to the predefined categories. We utilize  PointNet++ \cite{qi2017pointnet++} with multi-scale grouping as our backbone network to learn discriminating point-wise features for describing the raw points, which can be flexibly replaced by other 3D neural networks. Our network extracts a fixed number of \textit{N} points in one training sample. If \textit{NUM}, the number of points in the sample, is greater than \textit{N}, the points will be randomly selected. Otherwise, the points will be randomly repeated. Considering that \textit{NUM} lost during sampling has an impact on the neural network classification process, we define a feature $n =  \frac { NUM - N } { N }$ to compensate, which describes the relative relationship between \textit{NUM} and \textit{N}. PointRCNN \cite{shi2018pointrcnn} adds the distance information to the point feature, but we find that it has little effect in our experiments. Finally, we concatenate point coordinates $\left(x^{l}, y^{l}, z^{l}\right) \in \mathbb{R}^{3}$ in local coordinate system, normalized laser reflection intensity $i \epsilon[0,1]$ of the point measured by LiDAR and relative points number $n \in \mathbb{R}$ in cluster to a feature vector $
p=\left\{x^{l}, y^{l}, z^{l}, i, n\right\}$.

\section{Experiments} 

\subsection{Implementation details}\label{imple}
\smallskip
\noindent
{\bf 1) Dataset: }
Our dataset is converted from KITTI raw dataset \cite{Geiger2013IJRR}, which has 12,915 frames. Since that KITTI LiDAR scans can be temporally correlated if they are from the same sequence, we ensured that frames in the training set do not appear in testing sequences. Similar to SqueezeSeg \cite{wu2018squeezeseg}, we split the publicly available raw dataset\cite{Geiger2013IJRR} into a training set with 9,234 frames and a testing set with 3,681 frames. We evaluate on `Car', `Pedestrian', `Cyclist' categories and set `Van', `Truck', `Tram', `Person$\_$sitting', `Misc' and `DontCare' categories in KITTI as background class. Our training/testing split will be released as well. Some background samples are randomly discarded during our training due to the extremely unbalanced data (the number of background samples is much higher than the others).

\smallskip
\noindent
{\bf 2) Cluster proposal:}
In our experiments, the parameters for the ground plane fitting are set as $N_{seg} =3$, $N_{iter} =3$, $N_{LPR} = 20$, $Th_{seeds} = 0.4m$ and $Th_{dist} = 0.3m$. For ring-based clustering algorithm, the parameters are $Th_{ring} = 0.5m$ and $Th_{prop} = 1m$. For proposals refinement, we set $Th_{num} = 30$ and enlarge the 3D oriented bounding box by $0.1m$ in both $x$ and $y$ axis, $0.4m$ in $z$ axis.
\renewcommand\arraystretch{1.3}
\begin{table}
	\small 
	\begin{center}
		\scalebox{1.0}
		{
			\begin{tabular}{c||ccc|c}
				\hline
				\multirow{2}{*}{Method} & 			
				\multicolumn{3}{c|}{Class} &
				\multirow{2}{*}{Average}\\
				& Car & Pedestrian & Cyclist \\
				
				\hline
				SqueezeSeg \cite{wu2018squeezeseg} & 64.6 & 21.8 & 25.1 & 37.2 \\
				SqueezeSegv2 \cite{wu2018squeezesegv2} & {\bf73.2} & 27.8 & 33.6 & 44.9 \\
				PointSeg \cite{wang2018pointseg} & 67.4 & 19.2 & 32.7 & 39.8 \\			
				\hline
				
				Ours & 63.3 & 34.3 & 40.8 & {46.1} \\
				 
				Ours+DA & 63.8 & 36.6 & {50.5} & {50.3} \\
				Ours+DA+\textit{n} & 63.3 & {\bf44.3} & {\bf50.8} & {\bf52.8} \\
				Ours+DA+FL & 63.3 & {40.2} & 44.9 & 49.5 \\
				Ours+DA+\textit{n}+FL & 62.8 & 40.6 & 46.2 & {49.9} \\
				\hline
			\end{tabular}
		}
	\end{center}
	\caption{Semantic segmentation performance of PASS3D. \textbf{Ours} denotes our baseline model. \textbf{+DA} denotes using data augmentation method. \textbf{+\textit{n}} denotes using feature \textit{n}. \textbf{+FL} denotes using focal loss. }
	\label{tab:val}
	\vspace{-3mm}
\end{table}

\smallskip
\noindent
{\bf 3) Training settings:}
We follow the neural network structure of PointNet++ \cite{qi2017pointnet++} and set the parameters similar to Frustum-PointNet \cite{qi2018frustum}, where three set abstraction layer with multi-scale grouping are used to downsample points into groups with sizes 128, 32, 1. Three feature propagation layers are used to obtain per-point feature for point-wise semantic segmentation. At the top of the network, two fully connected layers and a dropout layer (we set 0.7 keep probability in dropout layer) are used for classification. Cross-entropy loss is minimized during training. Besides, we also try focal loss \cite{lin2017focal} in experiments and the classification loss has the following form:
\begin{equation}
F L\left(p_{t}\right)=-\alpha\left(1-p_{t}\right)^{\gamma} \log \left(p_{t}\right)
\end{equation}
where \textit{$p_{t}$} is the model's estimated probability. \textit{$\alpha$} and \textit{$\gamma$} are the parameters of the focal loss and we set them be \textit{$0.25$} and \textit{$2$} respectively.

\subsection{Evaluation metrics}\label{compare}
The evaluation precision, recall and IoU (intersection-over-union) are defined as follows:

\vspace{-2mm}
\begin{equation}\label{ort_loss}
\textit{Pr}_{c} = \frac { \left| P _ { c } \cap G _ { c } \right| } { \left| P _ { c } \right| },
\textit{Re}_{c} = \frac { \left| P _ { c } \cap G _ { c } \right| } { \left| G _ { c } \right| },
{ IoU } _ { c } = \frac { \left| P _ { c } \cap G _ { c } \right| } { \left| P _ { c } \cup G _ { c } \right| }\nonumber
\end{equation}
\vspace{-2mm}

\noindent
where $P_c$ and $G_c$ respectively denote the predicted and ground-truth point sets that belong to class-c. $\left| \cdot \right|$ denotes the cardinality of a set.

\subsection{Experimental results}\label{compare}

\smallskip
\noindent
{\bf 1) Point-wise semantic segmentation: }
Similar to \cite{ren2015faster}, we evaluate our model’s performance on class-level segmentation tasks by a point-wise comparison of the predicted results with ground-truth labels. We employ IoU as our evaluation metric. We compare our method with the state-of-art methods in Tab.\ref{tab:val} and some semantic segmentation results are shown in Fig.\ref{fig:test_vis}. 

As shown in Tab.\ref{tab:val}, our method outperforms previous state-of-the-art methods with remarkable margins in `Pedestrian' and `Cyclist' categories, raising $\textbf{16.5\%}$ and $\textbf{17.2\%}$ respectively. We develop the average IoU from ${44.9\%}$ of SqueezeSegv2\cite{wu2018squeezesegv2} to $\textbf{52.8\%}$. Our data augmentation method strikingly improves the performance of all classes. Notably, it raises `Cyclist' IoU of our baseline model from $\textbf{40.8\%}$ to $\textbf{50.5\%}$, which proves its efficacy on recognitions of non-rigid objects. The added feature $n$ improves the IoU of `Pedestrian' and `Cyclist'. The effect of focal loss in our experiments is uncertain. One of the possible reasons why the semantic segmentation performance of `Car' categories is worse than others is that our stage-1 clusters the adjacent cars and trees into one proposal, resulting in a missed detection.

\smallskip
\noindent
{\bf2) Evaluation of stage-1:}
The performance of our cluster proposal algorithm based on fast point cloud segmentation \cite{zermas2017fast} is evaluated by calculating the recall of point-wise foreground label. We put `Car', `Pedestrian' and `Cyclist' classes as foreground while other classes as background. With only proposing about 30 clusters per frame, our method achieves \textbf{89.5\%} point-wise recall in \textbf{5ms}. 
To some extent, we hold that we have achieved a considerable recall with less redundant proposals, which introduces less computation into the stage-2 neural network and increases the overall speed. We only pass about $5k$ points in the entire scene to the stage-2 neural network, while other methods\cite{shi2018pointrcnn},\cite{yan2018second} usually pass all points ($\approx 30k$) into the network. Besides, our point-wise semantic segmentation method can predict the possibility of the category of each point even when stage-1 algorithm clusters different categories of points into one proposal, which means our method is robust to the stage-1 algorithm.

 \section{Conclusion}
 
We presented a two-stage precise and accelerated semantic segmentation framework for 3D point cloud, which combines the efficiency of traditional geometric methods and the robustness of deep learning methods. The proposed accelerated cluster proposal algorithm at stage-1 clusters point cloud effectively and generates less redundant refined proposals of high quality, which is competitive to previous excellent methods. Moreover, we have proposed a novel approach to augment data that makes full use of the advantages of point clouds, leading to amazing promotion in the semantic segmentation of non-rigid objects like pedestrians and cyclists. The experiments prove that our method is efficient and accurate, obtaining impressive results. 

 \section{Acknowledgement}

This work is supported by the National Natural Science Foundation of China under Grant U1509210, 61836015. We want to thank Xingguang Zhong and Zhen Zhang for fruitful discussions.

\addtolength{\topmargin}{0.249cm}
\bibliographystyle{ieeetr}
\bibliography{PASS3D}

\begin{thebibliography}{10}

\bibitem{zermas2017fast}
D.~Zermas, I.~Izzat, and N.~Papanikolopoulos, ``Fast segmentation of 3d point
  clouds: A paradigm on lidar data for autonomous vehicle applications,'' in
  {\em 2017 IEEE International Conference on Robotics and Automation (ICRA)},
  pp.~5067--5073, IEEE, 2017.

\bibitem{bogoslavskyi17pfg}
I.~Bogoslavskyi and C.~Stachniss, ``Efficient online segmentation for sparse 3d
  laser scans,'' {\em PFG -- Journal of Photogrammetry, Remote Sensing and
  Geoinformation Science}, pp.~1--12, 2017.

\bibitem{yang2018pixor}
B.~Yang, W.~Luo, and R.~Urtasun, ``Pixor: Real-time 3d object detection from
  point clouds,'' in {\em Proceedings of the IEEE Conference on Computer Vision
  and Pattern Recognition}, pp.~7652--7660, 2018.

\bibitem{liang2018deep}
M.~Liang, B.~Yang, S.~Wang, and R.~Urtasun, ``Deep continuous fusion for
  multi-sensor 3d object detection,'' in {\em Proceedings of the European
  Conference on Computer Vision (ECCV)}, pp.~641--656, 2018.

\bibitem{wu2018squeezeseg}
B.~Wu, A.~Wan, X.~Yue, and K.~Keutzer, ``Squeezeseg: Convolutional neural nets
  with recurrent crf for real-time road-object segmentation from 3d lidar point
  cloud,'' in {\em 2018 IEEE International Conference on Robotics and
  Automation (ICRA)}, pp.~1887--1893, IEEE, 2018.

\bibitem{wu2018squeezesegv2}
B.~Wu, X.~Zhou, S.~Zhao, X.~Yue, and K.~Keutzer, ``Squeezesegv2: Improved model
  structure and unsupervised domain adaptation for road-object segmentation
  from a lidar point cloud,'' {\em arXiv preprint arXiv:1809.08495}, 2018.

\bibitem{wang2018pointseg}
Y.~Wang, T.~Shi, P.~Yun, L.~Tai, and M.~Liu, ``Pointseg: Real-time semantic
  segmentation based on 3d lidar point cloud,'' {\em arXiv preprint
  arXiv:1807.06288}, 2018.

\bibitem{qi2018frustum}
C.~R. Qi, W.~Liu, C.~Wu, H.~Su, and L.~J. Guibas, ``Frustum pointnets for 3d
  object detection from rgb-d data,'' in {\em Proceedings of the IEEE
  Conference on Computer Vision and Pattern Recognition}, pp.~918--927, 2018.

\bibitem{xu2018pointfusion}
D.~Xu, D.~Anguelov, and A.~Jain, ``Pointfusion: Deep sensor fusion for 3d
  bounding box estimation,'' in {\em Proceedings of the IEEE Conference on
  Computer Vision and Pattern Recognition}, pp.~244--253, 2018.

\bibitem{ku2018joint}
J.~Ku, M.~Mozifian, J.~Lee, A.~Harakeh, and S.~L. Waslander, ``Joint 3d
  proposal generation and object detection from view aggregation,'' in {\em
  2018 IEEE/RSJ International Conference on Intelligent Robots and Systems
  (IROS)}, pp.~1--8, IEEE, 2018.

\bibitem{shi2018pointrcnn}
S.~Shi, X.~Wang, and H.~Li, ``Pointrcnn: 3d object proposal generation and
  detection from point cloud,'' {\em arXiv preprint arXiv:1812.04244}, 2018.

\bibitem{qi2017pointnet}
C.~R. Qi, H.~Su, K.~Mo, and L.~J. Guibas, ``Pointnet: Deep learning on point
  sets for 3d classification and segmentation,'' in {\em Proceedings of the
  IEEE Conference on Computer Vision and Pattern Recognition}, pp.~652--660,
  2017.

\bibitem{Geiger2013IJRR}
A.~Geiger, P.~Lenz, C.~Stiller, and R.~Urtasun, ``Vision meets robotics: The
  kitti dataset,'' {\em International Journal of Robotics Research (IJRR)},
  2013.

\bibitem{douillard2011segmentation}
B.~Douillard, J.~Underwood, N.~Kuntz, V.~Vlaskine, A.~Quadros, P.~Morton, and
  A.~Frenkel, ``On the segmentation of 3d lidar point clouds,'' in {\em 2011
  IEEE International Conference on Robotics and Automation}, pp.~2798--2805,
  IEEE, 2011.

\bibitem{moosmann2009segmentation}
F.~Moosmann, O.~Pink, and C.~Stiller, ``Segmentation of 3d lidar data in
  non-flat urban environments using a local convexity criterion,'' in {\em 2009
  IEEE Intelligent Vehicles Symposium}, pp.~215--220, IEEE, 2009.

\bibitem{shin2017real}
M.-O. Shin, G.-M. Oh, S.-W. Kim, and S.-W. Seo, ``Real-time and accurate
  segmentation of 3-d point clouds based on gaussian process regression,'' {\em
  IEEE Transactions on Intelligent Transportation Systems}, vol.~18, no.~12,
  pp.~3363--3377, 2017.

\bibitem{wang2012could}
D.~Z. Wang, I.~Posner, and P.~Newman, ``What could move? finding cars,
  pedestrians and bicyclists in 3d laser data,'' in {\em 2012 IEEE
  International Conference on Robotics and Automation}, pp.~4038--4044, IEEE,
  2012.

\bibitem{chen2017multi}
X.~Chen, H.~Ma, J.~Wan, B.~Li, and T.~Xia, ``Multi-view 3d object detection
  network for autonomous driving,'' in {\em Proceedings of the IEEE Conference
  on Computer Vision and Pattern Recognition}, pp.~1907--1915, 2017.

\bibitem{qi2017pointnet++}
C.~R. Qi, L.~Yi, H.~Su, and L.~J. Guibas, ``Pointnet++: Deep hierarchical
  feature learning on point sets in a metric space,'' in {\em Advances in
  Neural Information Processing Systems}, pp.~5099--5108, 2017.

\bibitem{yan2018second}
Y.~Yan, Y.~Mao, and B.~Li, ``Second: Sparsely embedded convolutional
  detection,'' {\em Sensors}, vol.~18, no.~10, p.~3337, 2018.

\bibitem{zhou2018voxelnet}
Y.~Zhou and O.~Tuzel, ``Voxelnet: End-to-end learning for point cloud based 3d
  object detection,'' in {\em Proceedings of the IEEE Conference on Computer
  Vision and Pattern Recognition}, pp.~4490--4499, 2018.

\bibitem{zhai2019poseconvgru}
G.~Zhai, L.~Liu, L.~Zhang, and Y.~Liu, ``Poseconvgru: A monocular approach for
  visual ego-motion estimation by learning,'' {\em arXiv preprint
  arXiv:1906.08095}, 2019.

\bibitem{lin2017focal}
T.-Y. Lin, P.~Goyal, R.~Girshick, K.~He, and P.~Doll{\'a}r, ``Focal loss for
  dense object detection,'' in {\em Proceedings of the IEEE international
  conference on computer vision}, pp.~2980--2988, 2017.

\bibitem{ren2015faster}
S.~Ren, K.~He, R.~Girshick, and J.~Sun, ``Faster r-cnn: Towards real-time
  object detection with region proposal networks,'' in {\em Advances in neural
  information processing systems}, pp.~91--99, 2015.

\end{thebibliography}

\end{document}